%% file: camera_ready.tex
\pdfoutput=1

\documentclass[11pt]{article}
\usepackage{lipsum} 
\newcounter{para}

\usepackage[final]{acl}

\usepackage{times}
\usepackage{latexsym}
\usepackage{url}

\usepackage[T1]{fontenc}

\usepackage{tabularx}
\usepackage{booktabs}
\usepackage{multirow}
\usepackage{adjustbox}

\usepackage[utf8]{inputenc}

\usepackage{microtype}

\usepackage{inconsolata}

\usepackage{graphicx}
\usepackage{amsmath}
\usepackage{float}
\usepackage{subcaption}
\title{\felix{\textsc{AlignFreeze}: Navigating the Impact of Realignment on the Layers of Multilingual Models Across Diverse Languages}}

\author{
Steve Bakos\textsuperscript{\rm 1}~~~ Félix Gaschi\textsuperscript{\rm 3}~~~ David Guzmán\textsuperscript{\rm 2}\\ \bf Riddhi More\textsuperscript{\rm 1}~~~ Kelly Chutong Li\textsuperscript{\rm 2}~~~ En-Shiun Annie Lee\textsuperscript{\rm 1,2} \\
  \textsuperscript{\rm 1}Ontario Tech University, Canada~~~  \textsuperscript{\rm 2}University of Toronto, Canada\\
  \textsuperscript{\rm 3}SAS Posos, France\\
  \texttt{felix@posos.co}}

\newif\ifshowcomments
\newif\ifshowcommentsbis

\newif\ifshowrevision
\showrevisionfalse

\newcommand{\david}[1]{\ifshowcomments\textcolor{orange}{#1}\else#1\fi}

\newcommand{\steve}[1]{\ifshowcomments\textcolor{purple}{#1}\else#1\fi}
\newcommand{\felix}[1]{\ifshowcomments\textcolor{brown}{#1}\else#1\fi}
\newcommand{\felixbis}[1]{\ifshowcommentsbis\textcolor{teal}{#1}\else#1\fi}
\newcommand{\revision}[1]{\ifshowrevision{\color{orange} #1}\else#1\fi}

\begin{document}
\maketitle

\begin{abstract}
Realignment techniques are often employed to enhance cross-lingual transfer in multilingual language models, still, they can sometimes degrade performance in languages that differ significantly from the fine-tuned source language.
This paper introduces \textsc{AlignFreeze}, a method that freezes either the layers' lower half or upper half during realignment. Through controlled experiments on \revision{4} tasks, 3 models, and in 35 languages, we find that realignment affects all the layers but can be the most detrimental to the lower ones. Freezing the lower layers can prevent performance degradation. Particularly, \textsc{AlignFreeze} improves Part-of-Speech (PoS) tagging performances in languages where full realignment fails: with XLM-R, it provides improvements of more than one standard deviation in accuracy in seven more languages than full realignment.
\end{abstract}

\section{Introduction}

Multilingual Language Models (mLMs) like XLM-R \citep{conneau-etal-2020-unsupervised} or mBERT \citep{devlin-etal-2019-bert} can perform cross-lingual transfer \citep{pires-etal-2019-multilingual,wu-dredze-2019-beto}. Once fine-tuned on a specific task in English, these models perform well on that same task when evaluated in other languages. While this can be useful for languages where fine-tuning data might be missing, cross-lingual transfer is often less efficient for languages that differ greatly from English \citep{pires-etal-2019-multilingual}, which unfortunately are the languages that would benefit the most from such ability.

With an approach similar to building multilingual word embeddings \citep{lample2018word,zhang-etal-2017-earth,artetxe-etal-2018-robust},
realignment explicitly re-trains an mLM for multilingual alignment with the hope of improving its cross-lingual transfer abilities. While some work report some level of success \citep{Cao2020Multilingual,zhao-etal-2021-inducing,pan-etal-2021-multilingual,wang-etal-2019-cross}, systematic evaluations show that realignment does not consistently improve cross-lingual transfer abilities and can significantly degrade them in some cases \citep{Efimov_2023,wu-dredze-2020-explicit}.

The relative failure of realignment raises the question of whether better multilingual alignment necessarily implies stronger cross-lingual transfer abilities. Previous work has found that mLMs have good multilingual alignment, on top of their cross-lingual transfer abilities \citep{dou-neubig-2021-word,ebrahimi-etal-2023-meeting}, and there even seems to be a strong link between alignment and cross-lingual transfer \citep{gaschi-etal-2023-exploring}, although the correlation is not causation and it remains that realignment often fails.

\felix{If better alignment is linked to better cross-lingual transfer, we hypothesize that realignment has some adverse effect that induces catastrophic forgetting of other important features of the model.}

\felix{To better understand this side-effect of realignment and how the different layers are affected, we propose \textsc{AlignFreeze}. In this method, half of the model layers are frozen during realignment. With a simple controlled experiment, we compare the impact on the lower and the upper layers. We find that realignment impacts all layers, but is particularly detrimental on lower layers, namely for a low-level task like PoS tagging.}

\section{\felix{Background on realignment}}

\felix{Realignment explicitly enforces the multilingual alignment of embeddings produced by multilingual models. It trains a multilingual model to produce similar representations for corresponding words in translated sentences. Two resources are needed: a translation dataset and a word alignment tool which, in our experiments, is either FastAlign \citep{dyer-etal-2013-simple}, AwesomeAlign \citep{dou-neubig-2021-word}, or a simple look-up table based on bilingual dictionaries \citep{lample2018word} as proposed in \citet{gaschi-etal-2023-exploring}.}

\felix{In our experiments, we use the realignment method proposed by \citet{wu-dredze-2020-explicit}, where a contrastive loss maximizes the similarity between the
representations of a pair of corresponding words ($h$ and $\text{aligned}(h)$) compared to all other possible pairs of words in a batch ($\mathcal{H}$ of size $B$) of pairs of translated sentences:}

{
\small
\begin{equation}
\mathcal{L} (\theta) = \dfrac{1}{2B} \sum_{h\in \mathcal{H}} \log \dfrac{ \exp (\text{sim}(h, \text{aligned}(h) ) / T) } { \displaystyle\sum_{h'\in \mathcal{H}, h'\neq h} \exp (\text{sim}(h, h') / T) }
\end{equation}
}
\felix{$T$ is the temperature, a hyperparameter set to 0.1.}

\section{Methodology}

We introduce \textsc{AlignFreeze}, a realignment method that relies on partial freezing to preserve half of the weights of an mLM during realignment. Because full realignment was shown not to work consistently \citep{wu-dredze-2020-explicit}, we hypothesize that applying realignment on the whole model could trigger some catastrophic forgetting of information useful to downstream cross-lingual tasks. To help mitigate that and better understand the impact of realignment, \textsc{AlignFreeze} freezes half of the layers of the mLM during realignment only.

\paragraph{Freezing Strategies} For the sake of simplicity and to reduce the number of experimental runs, we work with only two freezing strategies: 1)  \textit{Front-freezing}, which freezes the lower-half layers while the remaining layers are realigned; and 2) \textit{Back-freezing}, which freezes upper-half layers instead. 

Assuming that basic linguistic features are encoded in the lower layers while the top ones retain higher-level information \citep{peters-etal-2018-deep}, \textit{Front-freezing} aims to preserve the foundational language understanding captured in the early layers while enabling task-specific adaptation in the later layers. \textit{Back-freezing} seeks to maintain the abstract, high-level representations developed in the deeper layers while fine-tuning the model's basic linguistic features. \david{Our approach intentionally employs a straightforward freezing strategy, not to establish a new state-of-the-art realignment method, but to better understand the conditions under which realignment fails and how to mitigate its failure.}

The freezing is applied only during realignment. Thus, \textsc{AlignFreeze} can be described with the following steps: 1) Take a multilingual Language Model (mLM), 2) Freeze half of its layers, 3) train the remaining weights for the realignment loss, 4) unfreeze the frozen layers, 5) perform fine-tuning on the whole model for cross-lingual transfer.

\section{Experiment Setup}

\begin{table}[!htp]
\centering
\renewcommand{\arraystretch}{1.1} 
\resizebox{\linewidth}{!}{%
\begin{tabular}{cll}
\toprule
                   & \textbf{Parameters}             & \textbf{Values}                \\ \midrule
\multirow{3}{*}[-0em]{\rotatebox[origin=c]{90}{\footnotesize\textsc{AlignFreeze}}} & \textbf{Freezing Strategies}    & no freezing (full), Front Half, Back Half \\
\cline{2-3}
 &
  \textbf{Word Alignment Methods} &
  \begin{tabular}[c]{@{}l@{}}FastAlign \citep{dyer-etal-2013-simple}, \\ AwesomeAlign \citep{dou-neubig-2021-word}, \\ Bilingual Dictionaries \citep{lample2018word}\end{tabular} \\ \hline
\multirow{3}{*}[-0.1em]{\begin{tabular}[c]{@{}c@{}}\rotatebox[origin=c]{90}{\footnotesize\textsc{Settings}}\end{tabular}} & \textbf{Tasks}                  & PoS tagging (34 lang.), \revision{NER (34 lang.)}, NLI (12 lang.)               \\
\cline{2-3}
                   & \textbf{Datasets}               & UD-PoS, \revision{NER}, XNLI                  \\
\cline{2-3}
                   & \textbf{Baseline Models}        & XLM-R, DistilMBERT             \\ \bottomrule
\end{tabular}%
}
\caption{Summary of the experimental setting.}
\label{tab:methods-strategies}
\end{table}

\paragraph{Datasets} \textit{Realignment Dataset}: We use the OPUS-100 dataset \cite{zhang-etal-2020-improving} for the realignment phase. OPUS-100 is a multilingual parallel corpus that includes sentence pairs across multiple languages.\\
\textit{Downstream Task Dataset}: We evaluate multilingual models on three tasks: PoS tagging, \revision{Named Entity Recognition (NER),} Natural Language Inference (NLI), \revision{and Question Answering (QA)}. For PoS tagging, we use the Universal Dependencies dataset \citep{daniel2020universal}, which provides annotated treebanks for a wide range of languages. \revision{For NER, we use the WikiANN dataset \citep{rahimi-etal-2019-massively}.} For NLI, we use the Cross-lingual Natural Language inference (XNLI) corpus \cite{conneau-etal-2018-xnli}. \revision{For QA, we use the XQuAD dataset \citep{artetxe-etal-2020-cross}.}

\paragraph{Models}  Following \citet{gaschi-etal-2023-exploring}, we work with three models: DistilMBERT \cite{sanh2019distilbert}, mBERT \citep{devlin-etal-2019-bert}, and XLM-R Base \cite{conneau-etal-2020-unsupervised}. DistilMBERT is a smaller version of mBERT \cite{devlin-etal-2019-bert} obtained through distillation \citep{sanh2019distilbert}. DistilMBERT, mBERT, and XLM-R are all Transformer-based masked multilingual models.

\paragraph{Languages} We use English as the source language for fine-tuning. \felixbis{We evaluate on 34 languages for PoS-tagging \revision{and NER,}, 12 for NLI, \revision{and 11 for QA}. For realignment, we use the 34 available languages for PoS tagging\revision{, NER,}, NLI, and \revision{QA}. Using the same setting allows for comparison of results across tasks and also improves the outcome (cf. Appendix \ref{appendix:lang_num})}. We use all the languages that our resources allow: every language must be present in the translation dataset, the
bilingual dictionaries, and one of the downstream datasets. The full list can be found in the \autoref{sec:appendix}.

Further details about the implementation can be found in Appendix \ref{sec:exp_details} and in the source code\footnote{ANONYMIZED}.

\begin{figure*}[!h]
    \centering
    \begin{subfigure}[b]{0.8\linewidth}
        \includegraphics[width=\linewidth]{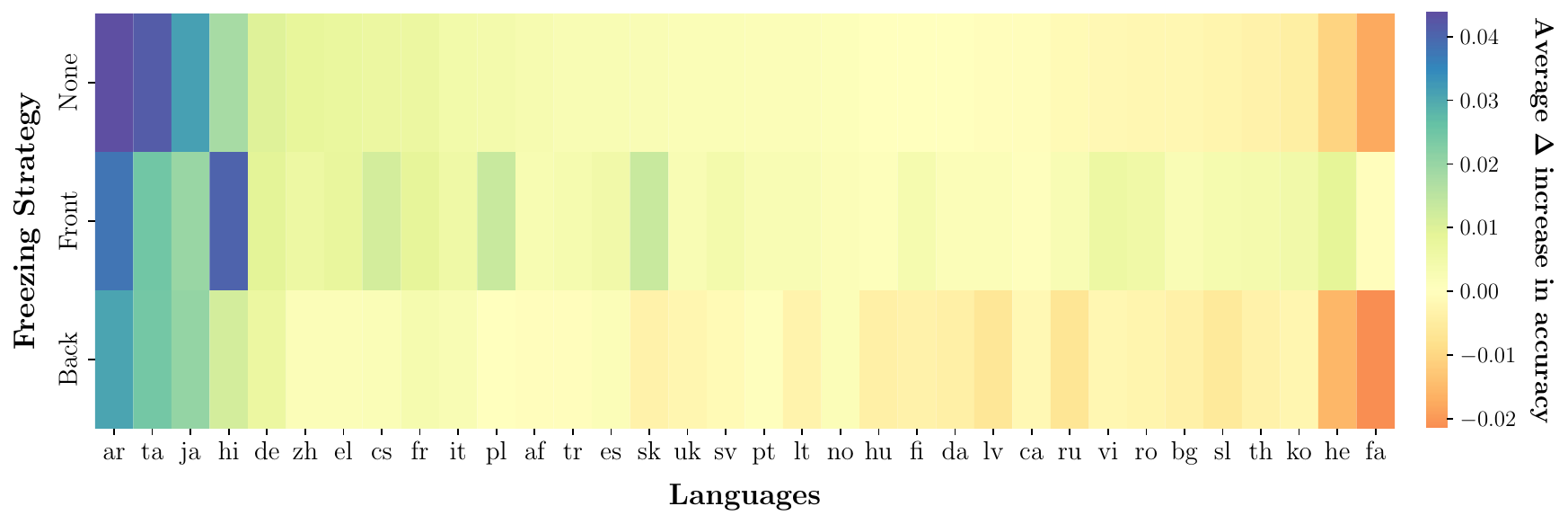}
        \caption{Variation of the accuracy with realignment with XLM-R Base for the PoS tagging task.}
        \label{fig:xlmr-udpos}
    \end{subfigure}
    
    \begin{subfigure}[b]{0.8\linewidth}
        \includegraphics[width=\linewidth]{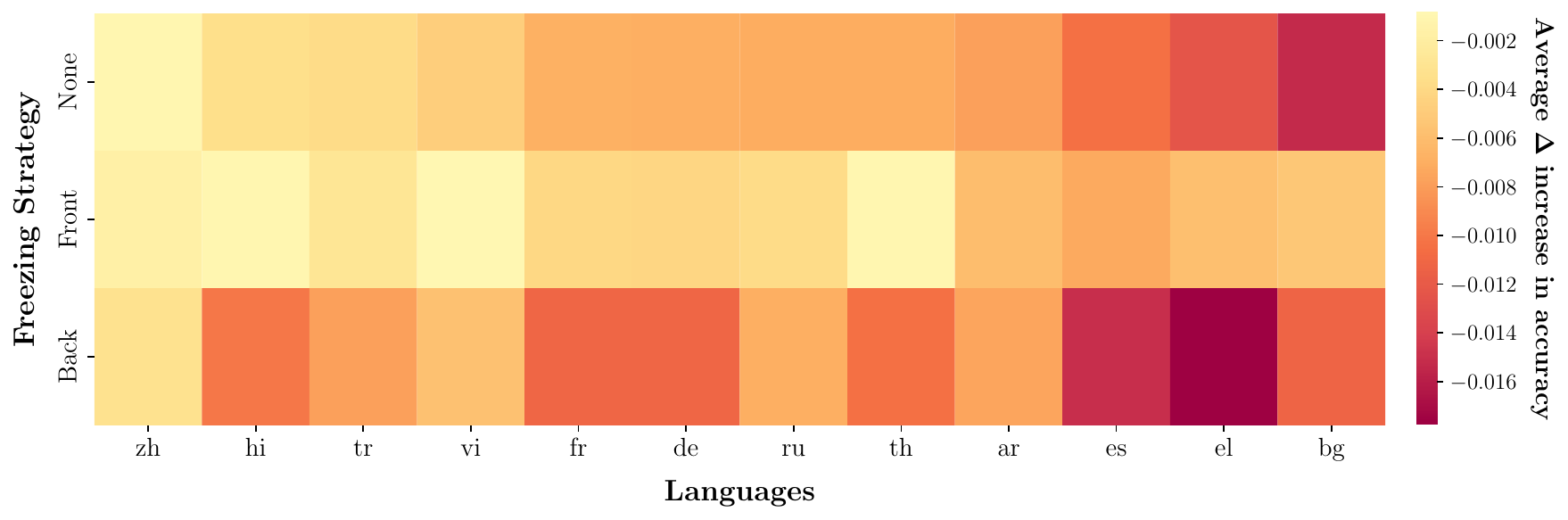}
        \caption{Variation of the accuracy with realignment with XLM-R Base for the NLI task.}
        \label{fig:xlmr-nli}
    \end{subfigure}
    
    \caption{Variation of the accuracies with realignment with XLM-R Base for the PoS tagging and NLI tasks. Languages are sorted by the improvement brought by full realignment. The average increase in accuracy is computed over 5 runs. Numerical values and results for other models can be found in Appendix \ref{appendix:additional_results}.}
    \label{fig:xlmr-heatmaps}
\end{figure*}

\section{Results and Discussion}


\paragraph{Finding 1: Full realignment fails in many cases.} 
\felix{As already observed by previous work \citep{wu-dredze-2020-explicit,Efimov_2023,gaschi-etal-2023-exploring}, full realignment isn't always successful. Table \ref{tab:avg_results} shows that realignment provides, on average, a significant improvement over fine-tuning with DistilMBERT, but the improvement is smaller with mBERT and even more so with XLM-R, especially for NLI \revision{and QA} where it even degrades the results.} \felix{Figure \ref{fig:xlmr-heatmaps} and Table \ref{tab:avg_results} also show that the outcome of full realignment varies a lot by language. For PoS-tagging with mBERT and distilMBERT, the majority of languages see a significant increase in accuracy. But with XLM-R, only 11 see a significant increase and one (Farsi) even undergoes a significant decrease of 2 points. For NLI, full realignment fails almost systematically with XLM-R, since 8 languages over 12 see a significant decrease in accuracy with realignment, \revision{while there can be as many significant increases and decreases for NER with XLM-R.}}

\begin{table*}[!h]
    \centering
    \adjustbox{max width=\linewidth}{
    \footnotesize
    \begin{tabular}{l|ccc|ccc|ccc|ccc|cc}
        \hline
     & \multicolumn{3}{c}{\textbf{PoS} (34 lang.)} & \multicolumn{3}{c}{\revision{\textbf{NER} (34 lang.)}} & \multicolumn{3}{c}{\textbf{NLI} (12 lang.)}  & \multicolumn{3}{c}{\revision{\textbf{QA} (11 lang.)}} & \multicolumn{2}{c}{\revision{Total (91)}} \\
         \cline{2-15}
         & acc. & \#$\downarrow$ & \#$\uparrow$ & acc. & \#$\downarrow$ & \#$\uparrow$ & acc. & \#$\downarrow$ & \#$\uparrow$ & F1 & \#$\downarrow$ & \#$\uparrow$ & \#$\downarrow$ & \#$\uparrow$ \\
         \hline
         \multicolumn{7}{l}{\textbf{\hspace{1.25em} DistilMBERT}}\\
         \hline
         Fine-tuning Only & 73.8$_{\pm 0.6}$ & - & - & 82.5$_{\pm 0.3}$ & - & - & 60.1$_{\pm 0.3}$ & - & - & 38.1$_{\pm 0.6}$ & -  &- & - & - \\
         Full realignment & 77.6$_{\pm 0.3}$ & 0 & 31 & 84.7$_{\pm 0.2}$ & 3 & 21 & 61.6$_{\pm 0.2}$ & 3 & 5 & 39.3$_{\pm 1.2}$ & 2 & 5 & 8 & 62 \\
         \textsc{AlignFreeze} (front) & 76.2$_{\pm 0.2}$ & 0 & 34 &  84.0$_{\pm 0.5}$ & 1 & 21 & 61.6$_{\pm 0.1}$ & 1 & 8 & 37.4$_{\pm 0.8}$ & 4 & 2 & 6 & 65\\
         \textsc{AlignFreeze} (back) & 77.4$_{\pm 0.1}$ & 0 & 30 &  83.7$_{\pm 0.7}$ & 4 & 17 & 61.9$_{\pm 0.2}$ & 1 & 6 & 39.1$_{\pm 1.0}$ & 2 & 5 & 7 & 58 \\
         \hline
         \multicolumn{7}{l}{\textbf{\hspace{1.25em} mBERT}}\\
         \hline
         Fine-tuning Only & 77.0$_{\pm 0.5}$ & - & - & 85.7$_{\pm 0.3}$ & - & - & 66.3$_{\pm 0.6}$ & - & - & 57.1$_{\pm 0.4}$ & - &- & - & - \\
         Full realignment &  79.6$_{\pm 0.4}$ & 1 & 32 & 86.4$_{\pm 0.3}$ & 19 & 4 &  67.4$_{\pm 0.4}$  & 0 & 8 & 52.9$_{\pm 0.7}$ & 11 & 0 &31 & 44  \\
         \textsc{AlignFreeze} (front) & 79.2$_{\pm 0.2}$  & 0 & 32  & \textbf{86.7$_{\pm 0.2}$} & 1 & 6 & 67.7$_{\pm 0.2}$ & 0 & 10 & 55.3$_{\pm 0.7}$ & 9 & 0 &  10 & 48\\
         \textsc{AlignFreeze} (back) & 79.3$_{\pm 0.3}$ & 1 & 30 &  86.5$_{\pm 0.6}$ & 12 & 6 & 67.5$_{\pm 0.3}$ & 0 & 10 & 53.7$_{\pm 0.6}$ & 11 & 0 & 24 & 46\\
         \hline 
         \multicolumn{7}{l}{\textbf{\hspace{1.25em} XLM-R Base}}\\
         \hline
         Fine-tuning Only & 80.9$_{\pm 0.1}$ & - & - & 84.9$_{\pm 0.4}$ & - & -  & \textbf{73.9}$_{\pm 0.2}$ & - & - & \textbf{61.2$_{\pm 0.4}$} & - & - & - & -\\
         Full realignment & 81.3$_{\pm 0.1}$ & 1 & 11 & 85.3$_{\pm 0.2}$ & 8 & 8 & 73.2$_{\pm 0.2}$ & 8 & 0 & 59.4$_{\pm 0.7}$ & 10 & 0 & 27 & 19\\
         \textsc{AlignFreeze} (front) & \textbf{81.7}$_{\pm 0.2}$ & 0 & 18 & 84.8$_{\pm 0.3}$ & 11 & 4 & 73.6$_{\pm 0.2}$ & 6 & 0 & 59.1$_{\pm 0.5}$ & 10 & 0 & 27 & 22 \\
         \textsc{AlignFreeze} (back) & 80.9$_{\pm 0.2}$ & 7 & 4 & 84.9$_{\pm 0.1}$ & 13 & 7 & 72.9$_{\pm 0.3}$ & 11 & 0 & 58.0$_{\pm 1.1}$ & 11 & 0 & 42 & 11 \\
                 \hline
        
        \multicolumn{2}{l}{\revision{Total of \#$\downarrow$ and \#$\uparrow$  by task}} & \multicolumn{2}{c}{\revision{/102}}& & \multicolumn{2}{c}{\revision{/102}} & & \multicolumn{2}{c}{\revision{/36}} & & \multicolumn{2}{c}{\revision{/33}} & \multicolumn{2}{c}{\revision{/273}}\\
        \hline
         Full realignment & - & 2 & 74 & - & 30 & 33 & - & 11 & 13 & - & 6 & 6 & 64 & 125\\
         \textsc{AlignFreeze} (front) & - & 0 & 84 & - & 13 & 31 & - & 7 & 18 & - & 9 & 2 & 43 & 135\\
         \textsc{AlignFreeze} (back) & - & 8 & 64 & - & 29 & 30 & - & 12 & 16 & - & 11 & 10 & 73 & 115\\
                 \hline
    \end{tabular}
    }
    \caption{Average accuracy of all target languages for PoS tagging, \revision{NER}, and XNLI with all models and realignment approaches. \felix{The number of languages for which realignment provides an increase above one standard deviation is reported (\#$\uparrow$) as well as the number of languages for which it provides a decrease of more than one standard deviation (\#$\downarrow$), the remaining languages see no significant change.} The results shown are for the bilingual dictionary aligner. Results are averaged over five runs. $\pm$ indicates the standard deviation.}
    \label{tab:avg_results}
\end{table*}


\paragraph{Finding 2: \textsc{AlignFreeze} (front) mitigates some of the failures of realignment.} 
\felix{Freezing the lower layers during realignment often improves results for cases where full realignment fails. Table \ref{tab:avg_results} shows that it brings an average improvement over full realignment with XLM-R for PoS-tagging and NLI, with 0.4 percent increases for both, \revision{but not for NER or QA, although the standard deviation is higher for QA making the results less conclusive}. But more importantly, }\felix{for PoS tagging, all languages are positively or neutrally impacted by front-freezing. And with XLM-R, the improvement is significant for 7 more languages than full realignment.} On Figure \ref{fig:xlmr-heatmaps}, while Farsi (fa) and Hebrew (he) undergo a significant decrease with full realignment for PoS tagging, they do not with \textsc{AlignFreeze} and even benefit from a 1-point improvement in the case of Hebrew. There are other languages, like Slovakian (sk), Polish (pl), and Hindi (hi) where full realignment provides a smaller improvement than front-freezing. \revision{Similarly to PoS tagging, front-freezing with mBERT for NER reduces the number of languages that suffer from realignment (from 19 to 1), but this is not the case with XLM-R.} \felixbis{Contrary to PoS tagging \revision{and NER}, NLI \revision{and QA} do not benefit much from realignment, but front-freezing allows to reduce the number of languages for which realignment is detrimental \revision{for NLI}.}

\paragraph{Finding 3: Realignment impacts the entire model, but it seems detrimental to the lower layers while it can be beneficial to the upper ones.}
\felix{Front-freezing can mitigate some failure cases of full realignment, thus realignment can have a detrimental effect on the lower layers. On the other hand, back-freezing seems to have a less important impact on realignment. Table \ref{tab:avg_results} shows that back-freezing does not significantly improve over full realignment, and Figure \ref{fig:xlmr-heatmaps} suggests that it provides worse results than any other alignment method for PoS tagging and NLI. \revision{The only exception is QA, for which back-freezing seems to improve over full realignment for distilMBERT and mBERT, but this improvement is not significant compared to the high variance of the results.} This contradicts \citet{gaschi-etal-2023-exploring} who hypothesized that }since realignment appears to work better on smaller models, realignment might only have an impact on the upper layers of the model. Our results show that realignment impacts all layers and seems to be the most detrimental to the lower ones.

\subsection{Generalized Recommendations for Practitioners using \textsc{AlignFreeze}}

\paragraph{Full realignment should be used for smaller models and low-level tasks.} \felix{As already suggested by previous work \citep{gaschi-etal-2023-exploring}, full realignment works better for smaller models like DistilMBERT and} the technique proves beneficial for tasks involving lower-level linguistic features, as evidenced by more consistent improvements in PoS tagging, compared to NLI \revision{QA, or even NER} (Table \ref{tab:avg_results}). This finding is relevant for researchers and organizations facing computational constraints. \textsc{AlignFreeze} and full realignment enable the enhancement of smaller, resource-efficient models, achieving competitive results without large-scale models or extensive computational resources.

\paragraph{\textsc{AlignFreeze} improves upon full realignment for PoS-tagging.} \felix{Table \ref{tab:avg_results} shows that \textsc{AlignFreeze} is never detrimental to cross-lingual transfer and improves results for more languages than full realignment. For NLI, while \textsc{AlignFreeze} still provides better results than full realignment, it can still be detrimental to cross-lingual transfer in some languages.} This suggests \textsc{AlignFreeze} is most effective when applied to tasks relying on syntactic and morphological information preserved in the frozen layers.

\paragraph{Cross-lingual transfer is hard to predict}
The variability in effectiveness across languages, models, and tasks highlights the importance of tailored approaches in multilingual NLP. \felix{In a truly zero-shot context, it seems hard to determine the right method for cross-lingual transfer, as shown by our results and previous work \citep{schmidt-etal-2023-one,yarmohammadi-etal-2021-everything}.
If evaluation data is available in the target language, practitioners should try all methods available to improve cross-lingual transfer, as results vary a lot by setting.
}

\section{Conclusion}

\steve{This study introduces \textsc{AlignFreeze}, a method using partial freezing to improve cross-lingual transfer in multilingual language models. Our experiments demonstrate that \textsc{AlignFreeze} effectively mitigates the failure cases of partial realignment by preserving pre-trained knowledge in the lower layers.}

\felix{When it comes to cross-lingual transfer, there does not seem to be any "silver bullet" \citep{yarmohammadi-etal-2021-everything} method that works for all languages, models, and tasks. Like realignment itself, and other cross-lingual approaches, \textsc{AlignFreeze} can help for some situations but not others. \textsc{AlignFreeze} can at least be useful for cross-lingual PoS-tagging with XLM-R.}

\felix{\textsc{AlignFreeze} helps better understand how realignment works. It impacts all layers and can be most detrimental to the lower ones, which is more visible on low-level tasks like PoS-tagging, that might be encoded in lower layers \citep{peters-etal-2018-deep}. Realignment probably fails simply because it is applied to the whole model without hindrance, which explains \textsc{AlignFreeze} relative success but also the results of other methods based on adapters like MAD-X \citep{pfeiffer-etal-2020-mad}}.

\section{Ethics and Limitations}

\subsection{Limitations}

We worked with the languages available in the datasets we used, but this led to high-resource languages and European languages being over-represented. To evaluate the effectiveness of cross-lingual transfer and realignment, the accuracy was averaged over all languages for a given task and model. Using the average to analyze the results has its risks, as different sets of languages can then potentially lead to different conclusions. However, the average remains convenient for our analysis and it was completed with some language-wise analysis as in Figures \ref{fig:xlmr-nli} and \ref{fig:xlmr-udpos}. Moreover, detailed results are provided in Appendix \ref{appendix:full_results} for the interested reader.

The experiments of this paper could be extended to more tasks and more models. PoS tagging\revision{, NER,}, NLI, and \revision{QA} were chosen for their differences. PoS tagging is a more low-level task looking at word categories while NLI deals with understanding. Moreover, partial realignment works well for PoS tagging, whereas it provides weaker results with NLI \citep{gaschi-etal-2023-exploring}. \revision{NER is chosen to complement this analysis with a task that is word-level, like PoS tagging, and semantic, like NLI. QA is chosen because it is a more difficult semantic tasks, like NLI, but is also a word-level one, like NER and PoS-tagging.} The choice of model was based on a similar approach. XLM-R Base is the largest mLM that we could train with our experimental setting while DistilMBERT offered a smaller alternative, and mBERT some middle ground. XLM-R was shown not to benefit too much from realignment, while DistilMBERT observes a large performance increase and can sometimes match XLM-R with the help of realignment \citep{gaschi-etal-2023-exploring}.

Throughout this paper, realignment is applied to encoder-only Language Models like DistilMBERT or XLM-R. While the literature on realignment also focuses on encoders \citep{Cao2020Multilingual,zhao-etal-2021-inducing,Efimov_2023,wu-dredze-2020-explicit}, realignment could be extended to more recent decoder-only generative multilingual models like Bloom \citep{workshop2023bloom} or XGLM \citep{lin-etal-2022-shot}. However, these models are often intended to be used in a zero-shot or few-shot fashion, and \citet{ahuja-etal-2023-mega} showed that cross-lingual transfer with fine-tuning of XLM-R largely outperforms prompt-based approaches with generative models on classification tasks.

\revision{This study experiments only with two simple freezing strategies: front-freezing and back-freezing. More granular freezing strategies could be designed to better understand the role of each layer. However, we experimented with several other approaches, but the results were not conclusive enough to include in the paper. Freezing half of the model does influence realignment, though the overall impact is already relatively minor. More granular freezing strategies led to even smaller variations (See Appendix \ref{sec:more_granular} for some results).} 

\revision{Some languages seem to benefit more from realignment than others. This study shows that freezing the bottom half of the layers during realignment might help with some languages that do not benefit from full realignment. However, \textsc{AlignFreeze}, like full realignment, does not work for all languages, and it is still hard to determine in advance which language will benefit or not from realignment. This issue can be explored through a regression analysis of our realignment results, but the regressor we trained overfitted on language-specific features and wasn't generalizing across languages, which defeats its purpose (cf. Appendix \ref{sec:regression}). Further research is needed to better understand what makes realignment fail under some conditions and succeed in others, but it might need larger-scale experiments to get conclusive results.}

\subsection{Ethics statement}

The resources we relied on limited our choice of languages. While working with 35 languages in total, this work contributes to the overexposure of European languages in the scientific literature. However, our work demonstrates that realignment can have a very different impact depending on the language and proposes new ways to improve cross-lingual transfer. While our conclusions will not directly impact the speakers of low-resource languages, they pave the way for potentially useful applications.

\bibliography{anthology,custom_new}

\clearpage

\appendix

\section{Related Works}

Pre-trained multilingual language models have become the predominant approach for cross-lingual transfer tasks. Word alignment methods that depend on these models have also been proposed \citep{jalili-sabet-etal-2020-simalign, nagata-etal-2020-supervised}. Current realignment methods are typically applied to a multilingual pre-trained model before fine-tuning in a single language (usually English) and applying to other languages on tasks such as Natural Language Inference (NLI) \citep{conneau-etal-2018-xnli}, Named Entity Recognition (NER) \citep{rahimi-etal-2019-massively}, Part-of-speech tagging (PoS) \citep{daniel2020universal}, or Question Answering (QA) \citep{artetxe-etal-2020-cross}. This process is intended to enhance the model's ability to generalize to other languages for these tasks.

Realignment can be performed in different ways. \citet{Cao2020Multilingual} minimizes the l2 distance between translated pairs. But some regularization is needed to prevent the representations from collapsing, which can be done through an additional loss term \citep{Cao2020Multilingual,zhao-etal-2021-inducing}
or using contrastive learning \citep{wu-dredze-2020-explicit}. Since the alignment is done at the word level between contextualized representations, an alignment tool is needed to obtain translated pairs to realign. Most methods employ the statistical tool FastAlign \citep{dyer-etal-2013-simple}. However neural-based tools can be used like AwesomeAlign \citep{dou-neubig-2021-word}, which are indeed shown to work better for low-resource languages, although they come at a larger computational cost \citep{ebrahimi-etal-2023-meeting}. A bilingual dictionary can also be used as a look-up table but extracts fewer pairs of words \citep{gaschi-etal-2023-exploring}. Empirically, it was however shown that realignment has inconsistent results when evaluated across several tasks and languages \citep{Efimov_2023,wu-dredze-2020-explicit}.

The failure of realignment questions the very link between multilingual alignment and cross-lingual transfer \citep{gaschi2022multilingual}. Realignment can increase multilingual alignment, but it might also be detrimental to some monolingual or even multilingual features learned by the model. To alleviate this, \citet{gaschi-etal-2023-exploring} tried to optimize the realignment loss jointly with the fine-tuning loss, but they did not report improved performances.

Due to its black-box nature, it is not straightforward to determine what role each layer of an mLM plays, but \citet{peters-etal-2018-deep} empirically showed, for ELMo, that the lower layers might encapsulate more lower-level information like syntax while the top ones relate to semantics. In a multilingual setting, \citet{wu-dredze-2019-beto} showed that freezing the lower layers of mBERT during fine-tuning can increase its cross-lingual performances.

\section{Additional Experimental details}\label{sec:exp_details}

\subsection{Languages}
\label{sec:appendix}

For PoS tagging\revision{ and NER}, because we used languages that were available simultaneously in the dataset but also in the different resources used for that task (bilingual dictionaries and the translation dataset), we worked with the following 34 languages: Afrikaans, Arabic, Bulgarian, Catalan, Chinese, Czech, Danish, Finnish, French, German, Greek, Hebrew, Hindi, Hungarian, Italian, Japanese, Korean, Latvian, Lithuanian, Norwegian, Persian, Polish, Portuguese, Romanian, Russian, Slovak, Slovenian, Spanish, Swedish, Tamil, Thai, Turkish, Ukrainian, and Vietnamese.

For NLI, due to similar constraints, we worked with the following 12 languages: Arabic, Bulgarian, Chinese, French, German, Greek, Hindi, Russian, Spanish, Thai, Turkish, and Vietnamese.

\subsection{Model Settings}
\label{B}
For both experiments, we reused the experimental setup from \citet{gaschi-etal-2023-exploring}. All experiments were run with 5 random seeds and performed using Nvidia A40 GPUs.\\
We train up to 5 epochs for PoS-tagging \revision{and NER} and 2 epochs for NLI, with a learning rate of 2e-5, batch size of 32 for training and evaluation, and a maximum length of 200 for the source and target. For realignment, we use a maximum length of 96 and a batch size of 16.

\subsection{Word alignment tools}\label{appendix:word_alignment}

 We employ three word alignment methods: FastAlign \citep{dyer-etal-2013-simple}, AwesomeAlign \citep{dou-neubig-2021-word}, and Bilingual Dictionaries \citep{lample2018word}. From a translation dataset, pairs were extracted either using a bilingual dictionary, following \citet{gaschi2022multilingual}, with FastAlign or AwesomeAlign. For FastAlign, alignments were generated in both directions and then symmetrized using the grow-diag-final-and heuristic provided by FastAlign, following \citet{wu-dredze-2020-explicit}. In all extraction methods, only one-to-one alignments were retained, and trivial cases where both words were identical were discarded, also following \citet{wu-dredze-2020-explicit}.

We use the three aligners for PoS tagging, but only the bilingual dictionaries for NLI, \revision{QA, and NER}, because it takes longer to train on NLI than PoS tagging \revision{and to avoid performing too many unnecessary experiments}. The approach based on bilingual dictionaries is preferred, as it is the aligner that provided the best results in \citet{gaschi-etal-2023-exploring}. Ultimately, the main part of the paper only reports the results with the bilingual dictionary, results with other aligners for PoS tagging are left at the end of the Appendix for the interested reader but do not impact our conclusions.

\subsection{Statistics about the datasets used}

The size of the datasets used for training and evaluating are reported in Table \ref{tab:dataset-size}.

\begin{table}[]
    \centering
    \adjustbox{max width=\linewidth}{
    \begin{tabular}{lcccc}
         & PoS-tagging & NLI & NER & QA \\
         \hline
         train (en) & 12,570 & 392,702 & 20,029 & 288,132\\
         \hline
        Afrikaans & 425 & - & 1,002 & - \\
        Arabic & 856 & 5010 & 10,000 & 4,317 \\
        Bulgarian & 1,117 & 5010 & 10,005 \\
        Catalan & 1,863 & - & 10,001 & -\\
        Chinese & 501 & 5010 & 10,378 & 3,831 \\
        Czech & 10,163 & - & 10,001& -\\
        Danish & 565 & - & 10,000& -\\
        Finnish & 1,000 & - & 10,000 & -\\ 
        French & 416 & 5010 & 10,000 & -\\
        German & 977 & 5010 & 10,000 & 3,405\\
        Greek & 478 & 5010 & 10,001 & 7,035 \\
        Hebrew & 509 & - & 10,000 & -\\
        Hindi & 1,685 & 5010 & 1,000 & 5,195 \\
        Hungarian & 451  & - & 10,004& -\\
        Italian & 485  & - & 10,000 & -\\
        Japanese & 546  & - & 11,724 & -\\
        Korean & 989  & - & 10,002& -\\
        Latvian & 1,828  & - & 10,002 & -\\
        Lithuanian & 687  & - & 10,000 & -\\
        Norwegian & 1,939  & - & 10,000 & -\\
        Persian & 1,456  & - & 10,000 & -\\
        Polish & 2,218  & - & 10,018& -\\
        Portuguese & 1,208  & - & 10,002& -\\
        Romanian & 734  & - & 10,000 & 4,174 \\
        Russian & 612 & 5010 & 10,000 & 4,109 \\
        Slovak & 1,061 & - & 10,001& -\\
        Slovenian & 790  & - & 10,018& -\\
        Spanish & 429 & 5010 & 10,000 & 3,391 \\
        Swedish & 1,000  & - & 10,000 & -\\
        Tamil & 125  & - & 1,000 & -\\
        Thai & 1,031& 5010 & 13,125 & 11,093 \\
        Turkish & 1,000  & 5010 & 10,001 & 3,839\\
        Ukrainian & 915  & - & 10,000& - \\
        Vietnamese & 800 & 5010 & 10,000 & 3,550\\
        \hline
    \end{tabular}
    }
    \caption{Size of the datasets (in number of samples) in the Universal Dependencies, NLI, \revision{NER, and QA} tasks.}
    \label{tab:dataset-size}
\end{table}

\subsection{Scientific artefacts used}

Here is a list of the scientific artifacts used\footnote{It does not include all the resources that are leveraged by those artifacts like specific Python packages.}:

\begin{itemize}
    \item The code for realignment comes from \citet{gaschi-etal-2023-exploring} and has MIT License
    \item the weights of DistilMBERT \citep{sanh2019distilbert} have License Apache-2.0
    \item the weights of XLM-R Base \citep{conneau-etal-2020-unsupervised} have MIT License
    \item The OPUS-100 dataset \citep{zhang-etal-2020-improving} does not have a known license, but it is a filtering of the OPUS corpus \citep{4992de1b5fb34f3e9691772606b36edf} which is itself the compilation of many translation datasets which are, to the best of our knowledge, free to be redistributed.
    \item The Universal Dependencies dataset \citep{daniel2020universal} is also a compilation of several datasets, which all have, to the best of our knowledge, open-source licenses.
    \item The XNLI corpus \citep{conneau-etal-2018-xnli} has a dedicated license but is nevertheless freely available for "typical machine learning use", which is the case in this paper.
\revision{
    \item The WikiANN dataset \citep{rahimi-etal-2019-massively} doesn't have a known license to the best of our knowledge. It is thus assumed to be free to use.
}
\revision{
    \item The XQuAD dataset \citep{artetxe-etal-2020-cross} has a the License CC-BY-SA-4.0, which allows its usage.
}
    \item FastAlign \citep{dyer-etal-2013-simple} has Apache-2.0 license
    \item AWESOME-align \citep{dou-neubig-2021-word} has BSD 3-Clause License
    \item The bilingual dictionaries \citep{lample2018word} have an "Attribution-NonCommercial 4.0 International" license that allows non-commercial use as is the case here
\end{itemize}

The scientific artifacts were thus used consistently with the intended use, as all identified licenses are open-source or authorize non-commercial use.

We cannot guarantee that the data we use do not contain personally identifying information or offensive content. However, this paper is not redistributing the data in any way and is simply using it for experiments. Nevertheless, we looked at randomly sampled elements of our datasets to verify their relevance and did not find any offensive or identifying content.

\section{Additional Results}\label{appendix:additional_results}
\subsection{Filtering data does not improve results}

\begin{figure}[htp]
\centering
\includegraphics[width=\linewidth]{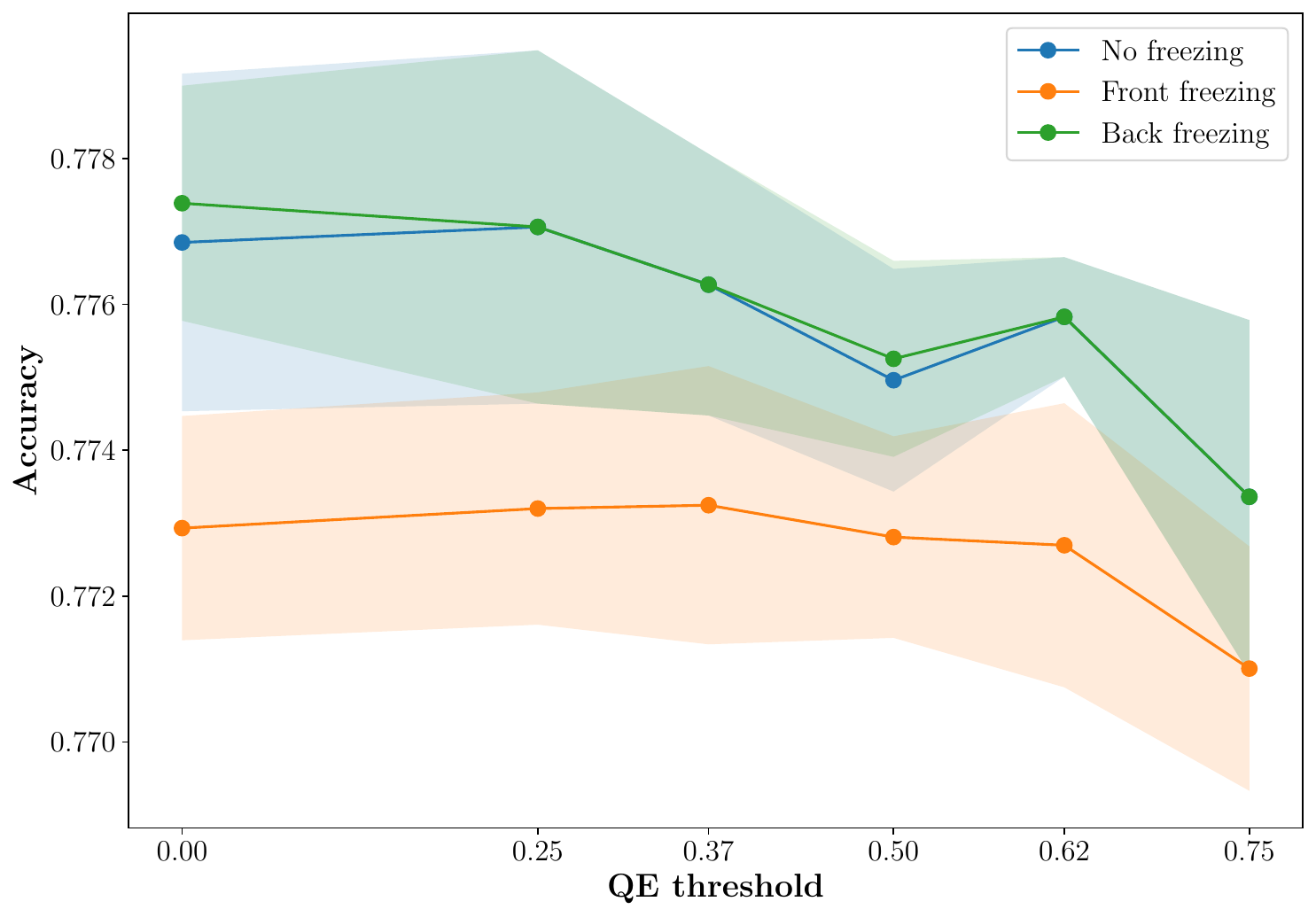}
\caption{Average accuracy for DistilMBERT when filtering the dataset for different percentiles of QE for the PoS tagging task.}
\label{fig:QE_filtering_accuracy}
\end{figure}

We hypothesized a direct correlation between the quality of the realignment results on the downstream tasks and the quality of the OPUS-100 dataset. To evaluate this, we employed a Quality Estimation (QE) model \citep{rei-etal-2022-cometkiwi} to selectively filter out sentence pairs below a predefined quality threshold. Since the OPUS-100 dataset contains significantly more sentences than needed for the realignment steps, the filtering should not affect the amount of data seen during realignment. Subsequently, we conducted experiments using this curated dataset to assess the impact of data quality on realignment results on the downstream tasks. Contrary to expectations, Figure \ref{fig:QE_filtering_accuracy} shows that, on average, using a higher quality dataset filtered by a QE model has little impact on the final results.

\subsection{Discussion on the amount of languages in realignment}\label{appendix:lang_num}

\felixbis{In this paper, realignment is performed with 34 languages for \revision{all tasks}, despite the downstream evaluation being possible in only 12 of those languages for NLI. In preliminary experiments, realignment was only performed on those 12 languages for NLI, and the whole set of 34 languages was used for PoS tagging \revision{and NER}. However, we eventually chose to use the same realignment step for both tasks, for a more controlled experiment, which means that we used 34 languages for NLI. As Table \ref{tab:lang_num} shows, realigning on 34 languages provides better results for all models except DistilMBERT.}

\felixbis{The evidence may be too anecdotal to conclude that using more languages for realignment generally provides better results. It might depend greatly on the alignment method used. Because we use an in-batch contrastive loss, adding languages increases diversity in the batch which might help the realignment work better. More extensive experiments in that regard are left for future work.}

\begin{table}[]
    \centering
    \adjustbox{max width=\linewidth}{
    \begin{tabular}{l|ccc|ccc}
        \hline
         & \multicolumn{3}{c}{12 languages} & \multicolumn{3}{c}{34 languages} \\
         \cline{2-7}
         & acc. & \#$\downarrow$ & \#$\uparrow$ & acc. & \#$\downarrow$ & \#$\uparrow$ \\
         \hline
         \multicolumn{7}{l}{\textbf{\hspace{1.25em} DistilMBERT}}\\
         \hline
         Fine-tuning Only & 60.1$_{\pm 0.3}$ & - & - & 60.1$_{\pm 0.3}$ & - & - \\
         Full realignment & 63.1$_{\pm 0.2}$ & 1 & 9 & 61.6$_{\pm 0.2}$ & 3 & 5\\
         \textsc{AlignFreeze} (front) & 62.7$_{\pm 0.3}$ & 0 & 11 & 61.6$_{\pm 0.1}$ & 1 & 8 \\
         \textsc{AlignFreeze} (back) & 63.1$_{\pm 0.2}$ & 0 & 10 & 61.9$_{\pm 0.2}$ & 1 & 6 \\
         \hline
         \multicolumn{7}{l}{\textbf{\hspace{2.25em} mBERT}}\\
         \hline
         Fine-tuning Only & 66.3$_{\pm 0.6}$ & - & - & 66.3$_{\pm 0.6}$ & - & - \\
         Full realignment &  66.9$_{\pm 0.7}$ & 0 & 4 &  67.4$_{\pm 0.4}$  & 0 & 8\\
         \textsc{AlignFreeze} (front) & 66.7$_{\pm 0.4}$ & 0 & 2 & 67.7$_{\pm 0.2}$ & 0 & 10\\
         \textsc{AlignFreeze} (back) & 67.0$_{\pm 0.7}$ & 0 & 4 & 67.5$_{\pm 0.3}$ & 0 & 10 \\
         \hline 
         \multicolumn{7}{l}{\textbf{\hspace{1.25em} XLM-R Base}}\\
         \hline
         Fine-tuning Only & \textbf{73.9}$_{\pm 0.2}$ & - & - & \textbf{73.9}$_{\pm 0.2}$ & - & - \\
         Full realignment & 72.9$_{\pm 0.1}$ & 11 & 0 & 73.2$_{\pm 0.2}$ & 8 & 0 \\
         \textsc{AlignFreeze} (front) & 73.4$_{\pm 0.1}$ & 9 & 0 & 73.6$_{\pm 0.2}$ & 6 & 0 \\
         \textsc{AlignFreeze} (back) & 73.2$_{\pm 0.3}$ & 11 & 0 & 72.9$_{\pm 0.3}$ & 11 & 0 \\
                 \hline

    \end{tabular}
    }
    \caption{Results of various realignment methods on NLI when using either 12 or 34 languages when performing realignment.}
    \label{tab:lang_num}
\end{table}

\revision{

\subsection{Additional results with more granular methods}\label{sec:more_granular}

Table \ref{tab:granular_strategy} shows the results of more granular strategies applied to PoS-tagging with DistilMBERT. While this combination and task and model is the one for which we observe the larger improvement with realignment, we do not observe any significantly interesting pattern for more granular freezing strategies. We tested two types of strategies: (1) freezing all layers except one during realignment (middle section of the table) and (2) freezing only one layer during realignment (bottom section of the table). While the first scenario shows some variation across layers, the number of languages that significantly benefit from these realignment strategies is lower than full realignment or front-freezing. For single-layer freezing, there isn't much variation across layers, and the results are very close to full realignment. This can be explained by the fact that by freezing only a single layer, we are not making as much as a difference from full realignment than when freezing half of the model.

\begin{table}[!h]
    \centering
    \adjustbox{max width=\linewidth}{
    \footnotesize
    \revision{
    \begin{tabular}{l|ccc}
        \hline
         & \multicolumn{3}{c}{\textbf{PoS-tagging} (34 lang.)}  \\
         \cline{2-4}
         & acc. & \#$\downarrow$ & \#$\uparrow$ \\
         \hline
         \multicolumn{4}{l}{\textbf{\hspace{1.25em} Baselines}}\\
         \hline
         Fine-tuning Only & 73.8$_{\pm 0.6}$ & - & -  \\
         Full realignment & 77.6$_{\pm 0.3}$ & 0 & 31 \\
         \textsc{AlignFreeze} (front) & 76.2$_{\pm 0.2}$ & 0 & \textbf{34} \\
         \textsc{AlignFreeze} (back) & 77.4$_{\pm 0.1}$ & 0 & 30 \\
         \hline
         \multicolumn{4}{l}{\textbf{\hspace{1.25em} Single-layer realignment}}\\
         \hline
         Layer 0 & 75.0$_{\pm 0.3}$ & 0 & 18 \\
         Layer 1 & 76.5$_{\pm 0.2}$ & 1 & 25 \\
         Layer 2 & 76.5$_{\pm 0.2}$ & 1 & 25 \\
         Layer 3 & 76.3$_{\pm 0.3}$ & 0 & 24 \\
         Layer 4 & 76.4$_{\pm 0.3}$ & 0 & 29 \\
         Layer 5 & 75.6$_{\pm 0.2}$ & 0 & 27 \\
         Layer 6 & 73.6$_{\pm 0.3}$ & 4 & 1 \\
         \hline 
         \multicolumn{4}{l}{\textbf{\hspace{1.25em} Single-layer freezing}}\\
         \hline
         Layer 0 & 77.7$_{\pm 0.2}$ & 0 & 30\\
         Layer 1 & 77.5$_{\pm 0.2}$ & 0 & 31 \\
         Layer 2 & 77.7$_{\pm 0.1}$ & 0 & 31\\
         Layer 3 & \textbf{77.8$_{\pm 0.2}$} & 0 & 32\\
         Layer 4 & 77.5$_{\pm 0.1}$ & 0 & 29 \\
         Layer 5 & 77.7$_{\pm 0.2}$ & 0 & 30\\
         Layer 6 & 77.7$_{\pm 0.2}$ & 0 & 30\\
                 \hline

    \end{tabular}
    }
    }
    \caption{\revision{Average accuracy of all target languages for PoS-tagging for distilMBERT with more granular freezing strategies. Refer to Table \ref{tab:avg_results} for more details on the notations.}}
    \label{tab:granular_strategy}
\end{table}

}

\revision{

\subsection{Realignment performance prediction}\label{sec:regression}

Some languages seem to benefit more than others from realignment. We performed a regression analysis using a random forest classifier to predict the ability to perform cross-lingual transfer from language-related and realignment-related features.

\paragraph{Prediction target}: the target variable for our regression model was the change in the model's accuracy with and without realignment for a given language. In other words, we compare the cross-lingual accuracy in a given language with and without realignment.

\paragraph{Input features}: as input features, we used various categorical features indicating the realignment method used: the aligner used (Fastalign, AWESOME-align, or bilingual dictionary), the freeze location (front or back freezing), and the freezing status (whether there is or isn't freezing). The language-related features are lang2vec distances from English \citep{littell2017uriel} (featural, syntactic, genetic, inventory, geographic, and phonological), word order, script type, and the language itself.

The random forest uses 30 estimators, with warm-start, bootstrapping, and the mean squared error as the splitting criterion. We perform the regression on the realignment results with Full realignment and \textsc{AlignFreeze} (front and back) for PoS-tagging with distilmBERT, because it is the configuration for which we have the higher variance in results and the larger amount of data points (all aligners were used). We also remove outliers using interquantile range method (IQR).

The fitted regressor has an $R^2$ score of 0.7126 and a mean squared error of 0.0001. The features' importance, aggregated by categories, is reported in Table \ref{tab:feature_importance}. While it seems that the lang2vec distances with English can largely help predict the effectiveness of realignment, this regression analysis has many limitations. First of all, while the $R^2$ score is adequate, attempts at generalizing the regressor to unseen languages provided poor results. The issue probably is that there aren't enough data points compared to the number of input features. The regressor overfits on language-related features because the language itself is a good predictor of the accuracy since results do not vary a lot across different seeds of realignment methods. 

In conclusion, realignment appears more effective for languages distant from English. However, since our regressor doesn't fully generalize to unseen languages, these findings should be interpreted with caution. We believe that additional data points are needed to draw more definitive conclusions, as the experiments in this paper provide a limited dataset.

\begin{table}[]
    \centering
    \revision{
    \begin{tabular}{l|c}
        feature &  importance \\
        \hline
        Lang2vec distance & 0.546 \\
        Language & 0.251 \\
        Script type & 0.077 \\
        Freeze location & 0.053 \\
        Aligner & 0.053 \\
        Freezing status & 0.011 \\
        Word order & 0.008
    \end{tabular}
    }
    \caption{\revision{Feature importance of various features of the random forest regressor applied to realignment results.}}
    \label{tab:feature_importance}
\end{table}

}

\subsection{Full Results}\label{appendix:full_results}

This section contains the detailed results of the experiments of this paper:

\begin{itemize}
    \item Realignment results for PoS tagging with DistilMBERT in Table \ref{table:results_distilMBERT_PoS-tagging_before}
\revision{
    \item Realignment results for NER with DistilMBERT in Table \ref{table:results_distilmBERT_NER_before}
}
    \item Realignment results for NLI with DistilMBERT in Table \ref{table:results_distilMBERT_XNLI_before}
\revision{
    \item Realignment results for QA with DistilMBERT in Table \ref{table:results_distilmBERT_xquad_before}
}
    \item Realignment results for PoS tagging with mBERT in table \ref{table:results_mBERT_PoS-tagging_before}
\revision{
    \item Realignment results for NER with mBERT in Table \ref{table:results_mBERT_NER_before}
}
    \item Realignment results for NLI with mBERT in table \ref{table:results_mBERT_xnli_before}
\revision{
    \item Realignment results for QA with mBERT in Table \ref{table:results_mBERT_xquad_before}
}
    \item Realignment results for PoS tagging with XLM-R in Table \ref{table:results_XLM-R_PoS-tagging_before}
\revision{
    \item Realignment results for NER with XLM-R in Table \ref{table:results_XLMR_NER_before}
}
    \item Realignment results for NLI with XLM-R in Table \ref{table:results_XLM-R_NLI_before}
\revision{
    \item Realignment results for QA with XLM-R in Table \ref{table:results_xlmr_xquad_before}
}
    \item Results of filtering for different percentiles of QE for NLI with DistilMBERT in Table \ref{table:results_distilMBERT_NLI_filtering}
    \item Results of filtering for different percentiles of QE for PoS tagging with DistilMBERT and FastAlign aligner in Table \ref{table:results_distilMBERT_PoS_filtering_fa}
    \item Results of filtering for different percentiles of QE for PoS tagging with DistilMBERT and AwesomeAlign aligner in Table \ref{table:results_distilMBERT_PoS_filtering_aa}
    \item Results of filtering for different percentiles of QE for PoS tagging with DistilMBERT and bilingual dictionary aligner in Table \ref{table:results_distilMBERT_PoS_filtering_bd}
\revision{
    \item Results of single-layer realignment for PoS tagging with DistilMBERT and bilingual dictionary aligner in Table \ref{table:results_mBERT_PoS-single_layer_realignment}
}
\revision{
    \item Results of single-layer freezing for PoS tagging with DistilMBERT and bilingual dictionary aligner in Table \ref{table:results_mBERT_PoS-single_layer_freezing}
}
\end{itemize}

\input{large_tables/pos-distilmbert}

\input{large_tables/ner-distilmbert}

\input{large_tables/nli-distilmbert}

\input{large_tables/xquad-distilmbert}

\input{large_tables/pos-mbert}

\input{large_tables/ner-mbert}

\input{large_tables/nli-mbert}

\input{large_tables/xquad-mbert}

\input{large_tables/pos-xlmr}

\input{large_tables/ner-xlmr}

\input{large_tables/nli-xlmr}

\input{large_tables/xquad-xlmr}
\input{large_tables/nli-distilmbert-filtering}
\input{large_tables/pos-distilmbert-fa-filtering}
\input{large_tables/pos-distilmbert-aa-filtering}
\input{large_tables/pos-distilmbert-bd-filtering}

\input{large_tables/pos-distilmbert-single-layer-realignment}

\input{large_tables/pos-distilmbert-single-layer-freezing}

\end{document}

%% file: large_tables/pos-distilmbert.tex
    \begin{table*}[ht]
    \centering
    \adjustbox{max width=\linewidth}{
\begin{tabular}{lcccccccccc}
    \toprule
     & \textbf{FT Only} & \multicolumn{3}{c}{\textbf{vanilla realignment}} & \multicolumn{3}{c}{\textbf{\textsc{AlignFreeze} with front-freezing}} & \multicolumn{3}{c}{\textbf{\textsc{AlignFreeze} with back-freezing}}\\ 
    \cline{2-11}
     & -  & FA & AA & BD & FA & AA & BD & FA & AA & BD \\
    \midrule
    Afrikaans & 85.5$_{\pm 0.2}$ & \textbf{86.4}$_{\pm 0.3}$ & \textbf{86.4}$_{\pm 0.3}$ & 85.6$_{\pm 0.4}$ & 86.2$_{\pm 0.2}$ & 86.3$_{\pm 0.3}$ & 86.1$_{\pm 0.3}$ & 86.0$_{\pm 0.2}$ & 86.0$_{\pm 0.3}$ & 85.4$_{\pm 0.1}$ \\
Arabic & 51.7$_{\pm 1.7}$ & 63.9$_{\pm 0.5}$ & 63.6$_{\pm 0.3}$ & \textbf{66.6}$_{\pm 0.5}$ & 63.3$_{\pm 0.5}$ & 63.0$_{\pm 0.5}$ & 65.0$_{\pm 0.6}$ & 63.5$_{\pm 0.6}$ & 62.8$_{\pm 0.7}$ & 65.3$_{\pm 0.3}$ \\
Bulgarian & 85.0$_{\pm 0.5}$ & 87.4$_{\pm 0.2}$ & \textbf{87.6}$_{\pm 0.3}$ & \textbf{87.6}$_{\pm 0.4}$ & 87.1$_{\pm 0.3}$ & 87.3$_{\pm 0.2}$ & 87.2$_{\pm 0.3}$ & 87.2$_{\pm 0.2}$ & \textbf{87.6}$_{\pm 0.2}$ & 87.5$_{\pm 0.2}$ \\
Catalan & 86.6$_{\pm 0.4}$ & 87.8$_{\pm 0.2}$ & 88.1$_{\pm 0.2}$ & \textbf{88.4}$_{\pm 0.1}$ & 87.6$_{\pm 0.3}$ & 87.8$_{\pm 0.2}$ & 88.2$_{\pm 0.1}$ & 87.9$_{\pm 0.2}$ & 88.2$_{\pm 0.2}$ & 88.1$_{\pm 0.2}$ \\
Chinese & 64.3$_{\pm 1.4}$ & 66.2$_{\pm 0.5}$ & 66.3$_{\pm 0.6}$ & \textbf{67.4}$_{\pm 0.7}$ & 66.6$_{\pm 0.5}$ & 66.3$_{\pm 0.4}$ & 67.3$_{\pm 0.6}$ & 66.2$_{\pm 0.7}$ & 66.3$_{\pm 0.7}$ & 66.7$_{\pm 0.5}$ \\
Czech & 79.1$_{\pm 0.7}$ & 84.6$_{\pm 0.3}$ & 84.7$_{\pm 0.4}$ & \textbf{85.3}$_{\pm 0.5}$ & 83.7$_{\pm 0.3}$ & 84.0$_{\pm 0.2}$ & 84.3$_{\pm 0.3}$ & 84.4$_{\pm 0.3}$ & 84.8$_{\pm 0.2}$ & 85.1$_{\pm 0.2}$ \\
\bottomrule
        \end{tabular}
        }
        \caption{PoS tagging average accuracy results across 5 seeds using DistilMBERT by freezing strategy, language, and aligner. Aligner names: FA - FastAlign, AA - AWESOME-align, BD - Bilingual Dictionary. The highest average accuracy value for each language is highlighted in bold.}
        \label{table:results_distilMBERT_PoS-tagging_before}
        \end{table*}

%% file: large_tables/ner-distilmbert.tex
\begin{table*}[ht]
\revision{
        \centering
        \adjustbox{max width=\linewidth}{
        \begin{tabular}{lcccc}
        \hline
        & \textbf{FT Only} & \multicolumn{1}{c}{\textbf{vanilla realignment}} & \multicolumn{1}{c}{\textbf{\textsc{AlignFreeze} with front-freezing}} & \multicolumn{1}{c}{\textbf{\textsc{AlignFreeze} with back-freezing}}\\ 
        \cline{2-5}
        & -  & BD & BD & BD \\
        \hline
    Afrikaans & 90.9$_{\pm 0.5}$ & \textbf{92.0}$_{\pm 0.3}$ & 91.8$_{\pm 0.3}$ & 91.9$_{\pm 0.4}$ \\
Arabic & 65.2$_{\pm 0.9}$ & 68.0$_{\pm 2.6}$ & 64.7$_{\pm 1.9}$ & \textbf{69.4}$_{\pm 2.7}$ \\
Bulgarian & 89.4$_{\pm 0.3}$ & \textbf{89.9}$_{\pm 0.2}$ & 89.7$_{\pm 0.2}$ & 89.6$_{\pm 0.4}$ \\
Catalan & 91.6$_{\pm 0.1}$ & \textbf{91.7}$_{\pm 0.0}$ & \textbf{91.7}$_{\pm 0.0}$ & 91.6$_{\pm 0.2}$ \\
Chinese & 76.8$_{\pm 0.4}$ & \textbf{78.1}$_{\pm 0.5}$ & 77.8$_{\pm 0.4}$ & 77.2$_{\pm 0.8}$ \\
Czech & 91.7$_{\pm 0.4}$ & \textbf{92.6}$_{\pm 0.2}$ & 92.4$_{\pm 0.2}$ & 92.4$_{\pm 0.1}$ \\
\hline
        \end{tabular}
        }
        \caption{\revision{NER accuracy results across 5 seeds using distilMBert by freezing strategy, language, and aligner. Aligner names: BD - Bilingual Dictionary. The highest average accuracy value for each language is highlighted in bold.}}
        \label{table:results_distilmBERT_NER_before}
    }
        \end{table*}

%% file: large_tables/nli-distilmbert.tex
        \begin{table*}[ht]
        \centering
        \adjustbox{max width=\linewidth}{
        \begin{tabular}{lcccc}
        \hline
        & \textbf{FT Only} & \multicolumn{1}{c}{\textbf{vanilla realignment}} & \multicolumn{1}{c}{\textbf{\textsc{AlignFreeze} with front-freezing}} & \multicolumn{1}{c}{\textbf{\textsc{AlignFreeze} with back-freezing}}\\ 
        \cline{2-5}
        & -  & BD & BD & BD \\
        \hline
    Arabic & 59.2$_{\pm 0.3}$ & 59.3$_{\pm 0.6}$ & \textbf{59.8}$_{\pm 0.4}$ & 59.2$_{\pm 0.5}$ \\
Bulgarian & 63.4$_{\pm 0.3}$ & 63.6$_{\pm 0.4}$ & \textbf{64.0}$_{\pm 0.2}$ & 63.8$_{\pm 0.5}$ \\
Chinese & 63.9$_{\pm 0.8}$ & 63.4$_{\pm 0.1}$ & \textbf{64.1}$_{\pm 0.5}$ & 63.4$_{\pm 0.5}$ \\
French & \textbf{70.1}$_{\pm 0.6}$ & 68.7$_{\pm 0.6}$ & 69.4$_{\pm 0.3}$ & 69.1$_{\pm 0.2}$ \\
German & 65.7$_{\pm 0.2}$ & 64.8$_{\pm 0.3}$ & \textbf{66.1}$_{\pm 0.5}$ & 65.5$_{\pm 0.6}$ \\
Greek & 60.8$_{\pm 0.4}$ & 62.0$_{\pm 0.9}$ & \textbf{62.9}$_{\pm 0.5}$ & 61.6$_{\pm 0.5}$ \\
Hindi & 54.1$_{\pm 0.6}$ & 54.9$_{\pm 1.0}$ & 55.3$_{\pm 0.3}$ & \textbf{55.6}$_{\pm 0.7}$ \\
Spanish & \textbf{70.0}$_{\pm 0.3}$ & 69.4$_{\pm 0.3}$ & 69.8$_{\pm 0.2}$ & \textbf{70.0}$_{\pm 0.3}$ \\
Thai & 36.1$_{\pm 0.5}$ & 47.1$_{\pm 1.7}$ & 42.0$_{\pm 1.4}$ & \textbf{47.4}$_{\pm 1.2}$ \\
Turkish & 57.0$_{\pm 0.5}$ & 58.7$_{\pm 0.5}$ & 58.1$_{\pm 0.6}$ & \textbf{58.7}$_{\pm 0.9}$ \\
Vietnamese & 57.6$_{\pm 2.3}$ & 64.3$_{\pm 0.3}$ & 63.9$_{\pm 0.7}$ & \textbf{65.0}$_{\pm 0.6}$ \\
\hline
Average & 60.1$_{\pm 0.2}$ & 61.6$_{\pm 0.2}$ & 61.6$_{\pm 0.1}$ & \textbf{61.9}$_{\pm 0.2}$ \\
        \hline
        \end{tabular}
        }
        \caption{XNLI average accuracy results across 5 seeds using DistilMBERT by freezing strategy, language, and aligner. Aligner names: BD - Bilingual Dictionary. The highest average accuracy value for each language is highlighted in bold.}
        \label{table:results_distilMBERT_XNLI_before}
        \end{table*}

%% file: large_tables/xquad-distilmbert.tex
\begin{table*}[ht]
        \centering
        \adjustbox{max width=\linewidth}{
        \revision{
        \begin{tabular}{lcccc}
        \hline
        & \textbf{FT Only} & \multicolumn{1}{c}{\textbf{vanilla realignment}} & \multicolumn{1}{c}{\textbf{\textsc{AlignFreeze} with front-freezing}} & \multicolumn{1}{c}{\textbf{\textsc{AlignFreeze} with back-freezing}}\\ 
        \cline{2-5}
        & -  & BD & BD & BD \\
        \hline
    Arabic & 37.4$_{\pm 0.7}$ & 38.1$_{\pm 1.2}$ & 38.3$_{\pm 1.2}$ & \textbf{38.8}$_{\pm 1.2}$ \\
Chinese & 35.7$_{\pm 0.9}$ & 36.8$_{\pm 1.3}$ & 36.3$_{\pm 1.6}$ & \textbf{38.3}$_{\pm 1.7}$ \\
German & 49.5$_{\pm 1.6}$ & 49.8$_{\pm 1.5}$ & 49.9$_{\pm 1.4}$ & \textbf{51.0}$_{\pm 1.0}$ \\
Greek & 32.4$_{\pm 1.0}$ & 33.9$_{\pm 1.5}$ & 33.4$_{\pm 0.7}$ & \textbf{34.9}$_{\pm 1.6}$ \\
Hindi & 29.4$_{\pm 0.9}$ & 29.6$_{\pm 0.8}$ & 30.1$_{\pm 0.4}$ & \textbf{30.2}$_{\pm 0.8}$ \\
Romanian & 44.2$_{\pm 1.9}$ & 46.4$_{\pm 2.4}$ & 44.9$_{\pm 2.0}$ & \textbf{47.3}$_{\pm 1.2}$ \\
Russian & 49.0$_{\pm 1.7}$ & 50.2$_{\pm 2.0}$ & 49.1$_{\pm 1.8}$ & \textbf{50.6}$_{\pm 1.8}$ \\
Spanish & 50.9$_{\pm 0.9}$ & 51.7$_{\pm 1.6}$ & 51.4$_{\pm 2.0}$ & \textbf{52.0}$_{\pm 1.5}$ \\
Thai & \textbf{18.7}$_{\pm 0.8}$ & 17.7$_{\pm 1.4}$ & 18.3$_{\pm 0.8}$ & 18.6$_{\pm 1.1}$ \\
Turkish & 31.0$_{\pm 0.5}$ & 32.8$_{\pm 1.2}$ & 32.1$_{\pm 0.5}$ & \textbf{33.3}$_{\pm 1.1}$ \\
Vietnamese & 38.0$_{\pm 0.5}$ & 41.3$_{\pm 2.7}$ & 38.4$_{\pm 1.3}$ & \textbf{41.5}$_{\pm 2.9}$ \\
\hline
Average & 37.8$_{\pm 0.6}$ & 38.9$_{\pm 1.1}$ & 38.4$_{\pm 0.9}$ & \textbf{39.7}$_{\pm 1.1}$ \\
        \hline
        \end{tabular}
        }
        }
        \caption{\revision{XQuAD average F1-score across 5 seeds using distilMBERT by freezing strategy, language, and aligner. Aligner names: BD - Bilingual Dictionary. The highest average accuracy value for each language is highlighted in bold.}}
        \label{table:results_distilmBERT_xquad_before}
        \end{table*}

%% file: large_tables/pos-mbert.tex
    \begin{table*}[ht]
    \centering
    \adjustbox{max width=\linewidth}{
    \begin{tabular}{lcccccccccc}
    \toprule
     & \textbf{FT Only} & \multicolumn{3}{c}{\textbf{vanilla realignment}} & \multicolumn{3}{c}{\textbf{\textsc{AlignFreeze} with front-freezing}} & \multicolumn{3}{c}{\textbf{\textsc{AlignFreeze} with back-freezing}}\\ 
    \cline{2-11}
     & -  & FA & AA & BD & FA & AA & BD & FA & AA & BD \\
    \midrule
Afrikaans & 87.0$_{\pm 0.4}$ & \textbf{88.4}$_{\pm 0.3}$ & 88.2$_{\pm 0.2}$ & 88.2$_{\pm 0.3}$ & 87.3$_{\pm 0.4}$ & 87.7$_{\pm 0.3}$ & 87.7$_{\pm 0.3}$ & 88.0$_{\pm 0.6}$ & 88.0$_{\pm 0.2}$ & 87.5$_{\pm 0.5}$ \\
Arabic & 51.0$_{\pm 0.5}$ & 63.7$_{\pm 1.6}$ & 63.9$_{\pm 1.0}$ & \textbf{65.1}$_{\pm 1.4}$ & 63.6$_{\pm 0.9}$ & 63.1$_{\pm 1.4}$ & 63.7$_{\pm 1.2}$ & 63.1$_{\pm 0.9}$ & 63.4$_{\pm 1.3}$ & 64.1$_{\pm 1.2}$ \\
Bulgarian & 86.3$_{\pm 0.8}$ & 87.9$_{\pm 0.7}$ & 88.1$_{\pm 0.3}$ & \textbf{88.1}$_{\pm 0.5}$ & 87.8$_{\pm 0.6}$ & 87.8$_{\pm 0.6}$ & 87.8$_{\pm 0.4}$ & 87.5$_{\pm 0.6}$ & 87.8$_{\pm 0.7}$ & 87.9$_{\pm 0.3}$ \\
Catalan & 86.7$_{\pm 0.3}$ & 88.2$_{\pm 0.3}$ & 88.6$_{\pm 0.3}$ & \textbf{89.0}$_{\pm 0.3}$ & 87.9$_{\pm 0.3}$ & 88.0$_{\pm 0.4}$ & 88.1$_{\pm 0.4}$ & 88.1$_{\pm 0.4}$ & 88.3$_{\pm 0.3}$ & 88.6$_{\pm 0.1}$ \\
Chinese & 65.7$_{\pm 1.0}$ & 67.9$_{\pm 1.3}$ & 67.4$_{\pm 0.1}$ & 69.0$_{\pm 0.5}$ & 67.6$_{\pm 1.1}$ & 66.8$_{\pm 0.1}$ & 69.0$_{\pm 0.8}$ & 68.4$_{\pm 1.1}$ & 68.2$_{\pm 0.6}$ & \textbf{69.7}$_{\pm 0.6}$ \\
Czech & 84.2$_{\pm 0.9}$ & 85.9$_{\pm 1.1}$ & 85.9$_{\pm 0.5}$ & \textbf{86.7}$_{\pm 0.5}$ & 86.1$_{\pm 0.8}$ & 86.0$_{\pm 0.8}$ & 86.4$_{\pm 0.4}$ & 85.6$_{\pm 0.8}$ & 85.7$_{\pm 0.9}$ & 86.5$_{\pm 0.5}$ \\
\bottomrule
    \end{tabular}
    }
    \caption{PoS tagging average accuracy results across 5 seeds using mBERT by freezing strategy, language, and aligner. Aligner names: FA - FastAlign, AA - AWESOME-align, BD - Bilingual Dictionary. The highest average accuracy value for each language is highlighted in bold.}
    \label{table:results_mBERT_PoS-tagging_before}
    \end{table*}

%% file: large_tables/ner-mbert.tex
\begin{table*}[ht]
        \centering
        \adjustbox{max width=\linewidth}{
        \revision{
        \begin{tabular}{lcccc}
        \hline
        & \textbf{FT Only} & \multicolumn{1}{c}{\textbf{vanilla realignment}} & \multicolumn{1}{c}{\textbf{\textsc{AlignFreeze} with front-freezing}} & \multicolumn{1}{c}{\textbf{\textsc{AlignFreeze} with back-freezing}}\\ 
        \cline{2-5}
        & -  & BD & BD & BD \\
        \hline
    Afrikaans & 92.8$_{\pm 0.2}$ & 92.6$_{\pm 0.5}$ & 92.7$_{\pm 0.2}$ & \textbf{92.8}$_{\pm 0.4}$ \\
Arabic & 67.1$_{\pm 0.9}$ & 68.9$_{\pm 1.5}$ & 68.9$_{\pm 1.8}$ & \textbf{70.7}$_{\pm 2.0}$ \\
Bulgarian & \textbf{90.7}$_{\pm 0.4}$ & 89.9$_{\pm 0.3}$ & 90.7$_{\pm 0.3}$ & 90.1$_{\pm 0.3}$ \\
Catalan & 92.8$_{\pm 0.2}$ & \textbf{92.9}$_{\pm 0.1}$ & 92.8$_{\pm 0.1}$ & 92.8$_{\pm 0.1}$ \\
Chinese & 78.9$_{\pm 0.7}$ & 78.7$_{\pm 0.6}$ & 79.0$_{\pm 0.6}$ & \textbf{79.6}$_{\pm 1.0}$ \\
Czech & 93.4$_{\pm 0.1}$ & 93.3$_{\pm 0.3}$ & \textbf{93.6}$_{\pm 0.1}$ & 93.2$_{\pm 0.1}$ \\
\hline
        \end{tabular}
        }
        }
        \caption{\revision{NER accuracy results across 5 seeds using mBERT by freezing strategy, language, and aligner. Aligner names: BD - Bilingual Dictionary. The highest average accuracy value for each language is highlighted in bold.}}
        \label{table:results_mBERT_NER_before}
        \end{table*}

%% file: large_tables/nli-mbert.tex
        \begin{table*}[ht]
        \centering
        \adjustbox{max width=\linewidth}{
        \begin{tabular}{lcccc}
        \hline
        & \textbf{FT Only} & \multicolumn{1}{c}{\textbf{vanilla realignment}} & \multicolumn{1}{c}{\textbf{\textsc{AlignFreeze} with front-freezing}} & \multicolumn{1}{c}{\textbf{\textsc{AlignFreeze} with back-freezing}}\\ 
        \cline{2-5}
        & -  & BD & BD & BD \\
        \hline
    Arabic & 64.6$_{\pm 0.5}$ & 65.0$_{\pm 0.6}$ & \textbf{65.6}$_{\pm 0.2}$ & 65.0$_{\pm 0.8}$ \\
Bulgarian & 68.0$_{\pm 0.8}$ & 69.1$_{\pm 0.6}$ & \textbf{69.3}$_{\pm 0.2}$ & 69.1$_{\pm 0.7}$ \\
Chinese & 68.9$_{\pm 0.6}$ & 69.5$_{\pm 0.7}$ & 69.2$_{\pm 0.4}$ & \textbf{69.9}$_{\pm 0.6}$ \\
French & 72.8$_{\pm 0.6}$ & 73.6$_{\pm 0.3}$ & \textbf{74.2}$_{\pm 0.3}$ & 73.7$_{\pm 0.5}$ \\
German & 70.1$_{\pm 0.5}$ & 70.3$_{\pm 0.6}$ & \textbf{71.0}$_{\pm 0.3}$ & 70.9$_{\pm 0.6}$ \\
Greek & 66.6$_{\pm 0.7}$ & 67.5$_{\pm 0.6}$ & \textbf{67.6}$_{\pm 0.6}$ & 67.4$_{\pm 0.8}$ \\
Hindi & 59.7$_{\pm 1.1}$ & 60.9$_{\pm 1.0}$ & \textbf{61.0}$_{\pm 0.5}$ & 61.0$_{\pm 0.3}$ \\
Spanish & 73.4$_{\pm 0.4}$ & 73.9$_{\pm 0.3}$ & \textbf{74.8}$_{\pm 0.3}$ & 74.2$_{\pm 0.3}$ \\
Thai & 53.3$_{\pm 2.3}$ & \textbf{57.4}$_{\pm 0.8}$ & 56.8$_{\pm 0.3}$ & 56.1$_{\pm 0.8}$ \\
Turkish & 61.4$_{\pm 0.5}$ & 63.5$_{\pm 0.6}$ & 63.2$_{\pm 0.4}$ & \textbf{63.8}$_{\pm 0.3}$ \\
Vietnamese & 69.0$_{\pm 0.5}$ & 70.3$_{\pm 0.2}$ & \textbf{70.9}$_{\pm 0.3}$ & 70.8$_{\pm 0.1}$ \\
\hline
Average & 66.3$_{\pm 0.6}$ & 67.4$_{\pm 0.4}$ & \textbf{67.7}$_{\pm 0.2}$ & 67.5$_{\pm 0.3}$ \\
        \hline
        \end{tabular}
        }
        \caption{XNLI average accuracy results across 5 seeds using mBERT by freezing strategy, language, and aligner. Aligner names: BD - Bilingual Dictionary. The highest average accuracy value for each language is highlighted in bold.}
        \label{table:results_mBERT_xnli_before}
        \end{table*}

%% file: large_tables/xquad-mbert.tex
\begin{table*}[ht]
        \centering
        \adjustbox{max width=\linewidth}{
        \revision{
        \begin{tabular}{lcccc}
        \hline
        & \textbf{FT Only} & \multicolumn{1}{c}{\textbf{vanilla realignment}} & \multicolumn{1}{c}{\textbf{\textsc{AlignFreeze} with front-freezing}} & \multicolumn{1}{c}{\textbf{\textsc{AlignFreeze} with back-freezing}}\\ 
        \cline{2-5}
        & -  & BD & BD & BD \\
        \hline
    Arabic & 55.5$_{\pm 1.2}$ & 54.6$_{\pm 0.6}$ & 55.0$_{\pm 1.0}$ & \textbf{55.6}$_{\pm 1.3}$ \\
Chinese & \textbf{53.1}$_{\pm 0.9}$ & 52.4$_{\pm 1.2}$ & 53.0$_{\pm 0.8}$ & 52.8$_{\pm 0.7}$ \\
German & 67.7$_{\pm 0.3}$ & 67.7$_{\pm 0.9}$ & 67.7$_{\pm 0.7}$ & \textbf{68.2}$_{\pm 0.2}$ \\
Greek & 53.0$_{\pm 0.9}$ & \textbf{53.5}$_{\pm 0.5}$ & 53.3$_{\pm 0.9}$ & 53.0$_{\pm 0.3}$ \\
Hindi & \textbf{49.1}$_{\pm 0.7}$ & 47.4$_{\pm 1.1}$ & 48.4$_{\pm 1.6}$ & 48.0$_{\pm 1.3}$ \\
Romanian & 66.1$_{\pm 0.6}$ & 66.8$_{\pm 0.2}$ & 66.6$_{\pm 0.6}$ & \textbf{66.9}$_{\pm 0.2}$ \\
Russian & \textbf{66.3}$_{\pm 0.8}$ & 64.9$_{\pm 0.2}$ & 65.1$_{\pm 0.6}$ & 65.6$_{\pm 0.4}$ \\
Spanish & 69.1$_{\pm 0.7}$ & 68.8$_{\pm 0.4}$ & 68.9$_{\pm 0.8}$ & \textbf{70.0}$_{\pm 0.6}$ \\
Thai & 35.5$_{\pm 0.9}$ & \textbf{35.6}$_{\pm 1.6}$ & 34.6$_{\pm 1.0}$ & 35.1$_{\pm 1.3}$ \\
Turkish & 47.2$_{\pm 1.4}$ & 46.8$_{\pm 1.4}$ & \textbf{47.6}$_{\pm 0.8}$ & 46.7$_{\pm 1.4}$ \\
Vietnamese & 63.7$_{\pm 0.6}$ & 62.9$_{\pm 0.9}$ & 63.7$_{\pm 0.6}$ & \textbf{63.7}$_{\pm 1.0}$ \\
\hline
Average & 56.9$_{\pm 0.3}$ & 56.5$_{\pm 0.5}$ & 56.7$_{\pm 0.5}$ & \textbf{56.9}$_{\pm 0.5}$ \\
        \hline
        \end{tabular}
        }
        }
        \caption{\revision{XQuAD average F1-score results across 5 seeds using mBERT by freezing strategy, language, and aligner. Aligner names: BD - Bilingual Dictionary. The highest average accuracy value for each language is highlighted in bold.}}
        \label{table:results_mBERT_xquad_before}
        \end{table*}

%% file: large_tables/pos-xlmr.tex
    \begin{table*}[ht]
    \centering
    \adjustbox{max width=\linewidth}{
    \begin{tabular}{lcccccccccc}
    \toprule
     & \textbf{FT Only} & \multicolumn{3}{c}{\textbf{vanilla realignment}} & \multicolumn{3}{c}{\textbf{\textsc{AlignFreeze} with front-freezing}} & \multicolumn{3}{c}{\textbf{\textsc{AlignFreeze} with back-freezing}}\\ 
    \cline{2-11}
     & -  & FA & AA & BD & FA & AA & BD & FA & AA & BD \\
    \midrule
    Afrikaans & 88.4$_{\pm 0.3}$ & 88.6$_{\pm 0.1}$ & 88.7$_{\pm 0.1}$ & \textbf{88.8}$_{\pm 0.1}$ & 88.6$_{\pm 0.2}$ & 88.6$_{\pm 0.2}$ & \textbf{88.8}$_{\pm 0.1}$ & 88.6$_{\pm 0.2}$ & 88.7$_{\pm 0.2}$ & 88.4$_{\pm 0.1}$ \\
Arabic & 63.2$_{\pm 0.8}$ & 65.5$_{\pm 0.9}$ & 65.3$_{\pm 1.1}$ & \textbf{67.6}$_{\pm 1.1}$ & 65.5$_{\pm 0.5}$ & 65.3$_{\pm 1.0}$ & 67.0$_{\pm 0.7}$ & 63.9$_{\pm 0.8}$ & 64.3$_{\pm 0.9}$ & 66.3$_{\pm 0.5}$ \\
Bulgarian & 89.3$_{\pm 0.5}$ & 89.1$_{\pm 0.3}$ & 89.4$_{\pm 0.2}$ & 89.1$_{\pm 0.2}$ & 89.5$_{\pm 0.1}$ & \textbf{89.9}$_{\pm 0.3}$ & 89.5$_{\pm 0.2}$ & 88.9$_{\pm 0.3}$ & 88.9$_{\pm 0.4}$ & 88.9$_{\pm 0.4}$ \\
Catalan & 89.4$_{\pm 0.5}$ & 89.2$_{\pm 0.2}$ & 89.5$_{\pm 0.3}$ & 89.4$_{\pm 0.4}$ & 89.6$_{\pm 0.5}$ & \textbf{89.8}$_{\pm 0.3}$ & 89.5$_{\pm 0.8}$ & 89.2$_{\pm 0.1}$ & 89.4$_{\pm 0.2}$ & 89.3$_{\pm 0.2}$ \\
Chinese & 71.4$_{\pm 0.4}$ & 70.5$_{\pm 0.4}$ & 70.7$_{\pm 0.8}$ & \textbf{72.2}$_{\pm 0.7}$ & 71.4$_{\pm 0.3}$ & 71.1$_{\pm 0.6}$ & 72.0$_{\pm 0.8}$ & 70.4$_{\pm 0.9}$ & 70.6$_{\pm 0.7}$ & 71.5$_{\pm 0.7}$ \\
Czech & 86.6$_{\pm 0.7}$ & 87.0$_{\pm 0.4}$ & 87.1$_{\pm 0.3}$ & 87.3$_{\pm 0.3}$ & 87.3$_{\pm 0.2}$ & 87.4$_{\pm 0.2}$ & \textbf{87.8}$_{\pm 0.3}$ & 86.6$_{\pm 0.7}$ & 86.7$_{\pm 0.6}$ & 86.8$_{\pm 0.7}$ \\
\bottomrule
        \end{tabular}
        }
        \caption{PoS tagging average accuracy results across 5 seeds using XLM-R by freezing strategy, language, and aligner. Aligner names: FA - FastAlign, AA - AWESOME-align, BD - Bilingual Dictionary. The highest average accuracy value for each language is highlighted in bold.}
        \label{table:results_XLM-R_PoS-tagging_before}
        \end{table*}

%% file: large_tables/ner-xlmr.tex
\begin{table*}[ht]
        \centering
        \adjustbox{max width=\linewidth}{
        \revision{
        \begin{tabular}{lcccc}
        \hline
        & \textbf{FT Only} & \multicolumn{1}{c}{\textbf{vanilla realignment}} & \multicolumn{1}{c}{\textbf{\textsc{AlignFreeze} with front-freezing}} & \multicolumn{1}{c}{\textbf{\textsc{AlignFreeze} with back-freezing}}\\ 
        \cline{2-5}
        & -  & BD & BD & BD \\
        \hline
    Afrikaans & 91.7$_{\pm 0.3}$ & 91.8$_{\pm 0.1}$ & 91.5$_{\pm 0.3}$ & \textbf{91.9}$_{\pm 0.3}$ \\
Arabic & 75.5$_{\pm 0.8}$ & \textbf{77.3}$_{\pm 1.1}$ & 76.4$_{\pm 1.3}$ & 76.8$_{\pm 1.4}$ \\
Bulgarian & 89.8$_{\pm 0.4}$ & 89.8$_{\pm 0.2}$ & 89.9$_{\pm 0.1}$ & \textbf{89.9}$_{\pm 0.3}$ \\
Catalan & \textbf{91.1}$_{\pm 0.1}$ & 90.9$_{\pm 0.2}$ & 91.0$_{\pm 0.1}$ & 90.8$_{\pm 0.1}$ \\
Chinese & \textbf{79.0}$_{\pm 0.8}$ & 78.4$_{\pm 0.4}$ & 78.4$_{\pm 0.6}$ & 78.0$_{\pm 0.3}$ \\
Czech & \textbf{92.5}$_{\pm 0.2}$ & 92.1$_{\pm 0.3}$ & 92.2$_{\pm 0.1}$ & 92.2$_{\pm 0.2}$ \\
\hline
        \end{tabular}
        }
        }
        \caption{\revision{NER accuracy results across 5 seeds using XLM-R Base by freezing strategy, language, and aligner. Aligner names: BD - Bilingual Dictionary. The highest average accuracy value for each language is highlighted in bold.}}
        \label{table:results_XLMR_NER_before}
        \end{table*}

%% file: large_tables/nli-xlmr.tex
    \begin{table*}[ht]
    \centering
    \adjustbox{max width=\linewidth}{
    \begin{tabular}{lcccccccccc}
    \toprule
     & \textbf{FT Only} & \multicolumn{1}{c}{\textbf{vanilla realignment}} & \multicolumn{1}{c}{\textbf{\textsc{AlignFreeze} with front-freezing}} & \multicolumn{1}{c}{\textbf{\textsc{AlignFreeze} with back-freezing}}\\ 
    \cline{2-5}
     & -  & BD & BD & BD \\
    \midrule
    Arabic & \textbf{70.8}$_{\pm 0.3}$ & 70.0$_{\pm 0.4}$ & 70.2$_{\pm 0.3}$ & 70.0$_{\pm 0.4}$ \\
Bulgarian & \textbf{77.0}$_{\pm 0.2}$ & 75.5$_{\pm 0.3}$ & 76.5$_{\pm 0.3}$ & 75.8$_{\pm 0.4}$ \\
Chinese & \textbf{72.5}$_{\pm 0.3}$ & 72.4$_{\pm 0.3}$ & 72.3$_{\pm 0.2}$ & 72.1$_{\pm 0.4}$ \\
French & \textbf{77.1}$_{\pm 0.1}$ & 76.4$_{\pm 0.3}$ & 76.7$_{\pm 0.1}$ & 76.0$_{\pm 0.3}$ \\
German & \textbf{75.7}$_{\pm 0.4}$ & 75.0$_{\pm 0.3}$ & 75.2$_{\pm 0.4}$ & 74.6$_{\pm 0.4}$ \\
Greek & \textbf{75.2}$_{\pm 0.3}$ & 74.0$_{\pm 0.3}$ & 74.7$_{\pm 0.2}$ & 73.5$_{\pm 0.4}$ \\
Hindi & \textbf{69.0}$_{\pm 0.3}$ & 68.7$_{\pm 0.5}$ & 68.9$_{\pm 0.5}$ & 68.0$_{\pm 0.8}$ \\
Spanish & \textbf{78.3}$_{\pm 0.2}$ & 77.3$_{\pm 0.2}$ & 77.6$_{\pm 0.2}$ & 76.8$_{\pm 0.2}$ \\
Thai & \textbf{71.1}$_{\pm 0.3}$ & 70.4$_{\pm 0.3}$ & 71.0$_{\pm 0.3}$ & 70.1$_{\pm 0.3}$ \\
Turkish & \textbf{71.9}$_{\pm 0.5}$ & 71.6$_{\pm 0.4}$ & 71.7$_{\pm 0.3}$ & 71.2$_{\pm 0.4}$ \\
Vietnamese & \textbf{73.8}$_{\pm 0.4}$ & 73.3$_{\pm 0.3}$ & 73.7$_{\pm 0.3}$ & 73.2$_{\pm 0.4}$ \\
\hline
Average & \textbf{73.9}$_{\pm 0.2}$ & 73.2$_{\pm 0.2}$ & 73.6$_{\pm 0.2}$ & 72.9$_{\pm 0.3}$ \\
        \bottomrule
        \end{tabular}
        }
        \caption{NLI average accuracy results across 5 seeds using XLM-R by freezing strategy, language, and aligner. Aligner names: BD - Bilingual Dictionary. The highest average accuracy value for each language is highlighted in bold.}
        \label{table:results_XLM-R_NLI_before}
        \end{table*}

%% file: large_tables/xquad-xlmr.tex
\begin{table*}[ht]
        \centering
        \adjustbox{max width=\linewidth}{
        \revision{
        \begin{tabular}{lcccc}
        \hline
        & \textbf{FT Only} & \multicolumn{1}{c}{\textbf{vanilla realignment}} & \multicolumn{1}{c}{\textbf{\textsc{AlignFreeze} with front-freezing}} & \multicolumn{1}{c}{\textbf{\textsc{AlignFreeze} with back-freezing}}\\ 
        \cline{2-5}
        & -  & BD & BD & BD \\
        \hline
    Arabic & \textbf{51.0}$_{\pm 1.0}$ & 50.8$_{\pm 0.4}$ & 49.9$_{\pm 0.7}$ & 50.8$_{\pm 0.3}$ \\
Chinese & \textbf{47.5}$_{\pm 0.8}$ & 47.1$_{\pm 0.6}$ & 46.4$_{\pm 0.6}$ & 46.6$_{\pm 0.8}$ \\
German & \textbf{65.7}$_{\pm 0.6}$ & 65.2$_{\pm 0.9}$ & 64.2$_{\pm 0.7}$ & 64.1$_{\pm 1.0}$ \\
Greek & \textbf{61.6}$_{\pm 0.7}$ & 60.9$_{\pm 0.9}$ & 58.8$_{\pm 0.8}$ & 59.4$_{\pm 0.7}$ \\
Hindi & \textbf{58.4}$_{\pm 1.0}$ & 57.7$_{\pm 0.7}$ & 56.2$_{\pm 0.7}$ & 56.4$_{\pm 0.8}$ \\
Romanian & \textbf{69.4}$_{\pm 0.5}$ & 68.9$_{\pm 0.8}$ & 67.7$_{\pm 1.0}$ & 68.2$_{\pm 0.5}$ \\
Russian & \textbf{66.3}$_{\pm 0.7}$ & 65.0$_{\pm 0.6}$ & 64.3$_{\pm 0.9}$ & 64.6$_{\pm 0.9}$ \\
Spanish & \textbf{67.8}$_{\pm 1.0}$ & 67.7$_{\pm 0.9}$ & 67.1$_{\pm 0.1}$ & 67.3$_{\pm 1.0}$ \\
Thai & \textbf{60.1}$_{\pm 0.9}$ & 57.9$_{\pm 2.1}$ & 56.9$_{\pm 1.2}$ & 57.6$_{\pm 0.4}$ \\
Turkish & \textbf{60.4}$_{\pm 0.7}$ & 60.3$_{\pm 1.1}$ & 59.6$_{\pm 0.5}$ & 60.0$_{\pm 0.6}$ \\
Vietnamese & \textbf{65.3}$_{\pm 0.2}$ & 65.2$_{\pm 0.4}$ & 64.5$_{\pm 0.7}$ & 64.4$_{\pm 0.9}$ \\
\hline
Average & \textbf{61.2}$_{\pm 0.4}$ & 60.6$_{\pm 0.6}$ & 59.6$_{\pm 0.5}$ & 59.9$_{\pm 0.4}$ \\
        \hline
        \end{tabular}
        }
        }
        \caption{\revision{XQuAD average F1-score results across 5 seeds using XLM-R Base by freezing strategy, language, and aligner. Aligner names: BD - Bilingual Dictionary. The highest average accuracy value for each language is highlighted in bold.}}
        \label{table:results_xlmr_xquad_before}
        \end{table*}

%% file: large_tables/nli-distilmbert-filtering.tex
    \begin{table*}[ht]
    \centering
    \adjustbox{max width=\linewidth}{
    \begin{tabular}{lccccccccccccc}
    \toprule
     & \textbf{FT Only} & \multicolumn{6}{c}{\textbf{vanilla realignment}} & \multicolumn{6}{c}{\textbf{\textsc{AlignFreeze} with front-freezing}}\\ 
    \cline{2-14}
     & -  & \multicolumn{2}{c}{\textbf{FA}} & \multicolumn{2}{c}{\textbf{AA}} & \multicolumn{2}{c}{\textbf{BD}} & \multicolumn{2}{c}{\textbf{FA}} & \multicolumn{2}{c}{\textbf{AA}} & \multicolumn{2}{c}{\textbf{BD}} \\
    \cline{2-14}
     & -  & 0\% & 50\% & 0\% & 50\% & 0\% & 50\% & 0\% & 50\% & 0\% & 50\% & 0\% & 50\%\\
    \midrule
    Arabic & 59.2$_{\pm 0.3}$ & 61.4$_{\pm 0.5}$ & 61.9$_{\pm 0.6}$ & \textbf{62.2}$_{\pm 0.4}$ & 61.6$_{\pm 0.3}$ & 60.2$_{\pm 0.6}$ & 59.6$_{\pm 1.3}$ & 60.2$_{\pm 0.7}$ & 60.7$_{\pm 0.4}$ & 60.9$_{\pm 0.4}$ & 61.0$_{\pm 0.4}$ & 60.6$_{\pm 0.5}$ & 60.0$_{\pm 0.1}$ \\
Bulgarian & 63.4$_{\pm 0.3}$ & 65.3$_{\pm 0.3}$ & 65.5$_{\pm 0.4}$ & \textbf{65.8}$_{\pm 0.4}$ & 65.4$_{\pm 0.5}$ & \textbf{65.8}$_{\pm 0.6}$ & 65.5$_{\pm 0.4}$ & 65.0$_{\pm 0.3}$ & 65.1$_{\pm 0.5}$ & 65.6$_{\pm 0.3}$ & 65.5$_{\pm 0.4}$ & 65.0$_{\pm 0.4}$ & 64.9$_{\pm 0.2}$ \\
Chinese & 63.9$_{\pm 0.9}$ & 65.1$_{\pm 0.6}$ & 65.4$_{\pm 0.6}$ & 64.9$_{\pm 0.4}$ & 64.6$_{\pm 0.3}$ & 64.3$_{\pm 0.1}$ & 63.9$_{\pm 0.6}$ & \textbf{65.5}$_{\pm 0.4}$ & \textbf{65.5}$_{\pm 0.7}$ & 65.4$_{\pm 0.3}$ & 65.2$_{\pm 0.3}$ & 65.4$_{\pm 0.5}$ & 65.0$_{\pm 0.5}$ \\
French & 70.1$_{\pm 0.7}$ & 69.3$_{\pm 0.3}$ & 69.5$_{\pm 0.2}$ & 70.0$_{\pm 0.4}$ & 69.6$_{\pm 0.6}$ & 69.0$_{\pm 0.4}$ & 69.9$_{\pm 0.4}$ & 69.7$_{\pm 0.5}$ & 69.9$_{\pm 0.3}$ & 70.1$_{\pm 0.1}$ & \textbf{70.5}$_{\pm 0.5}$ & 70.2$_{\pm 0.3}$ & 70.0$_{\pm 0.4}$ \\
German & 65.7$_{\pm 0.3}$ & 66.9$_{\pm 0.5}$ & 66.8$_{\pm 0.4}$ & 67.2$_{\pm 0.7}$ & 67.1$_{\pm 0.4}$ & 66.9$_{\pm 0.6}$ & 66.6$_{\pm 0.4}$ & \textbf{67.4}$_{\pm 0.3}$ & 67.1$_{\pm 0.5}$ & 67.1$_{\pm 0.5}$ & \textbf{67.4}$_{\pm 0.6}$ & 66.9$_{\pm 0.4}$ & 66.7$_{\pm 0.7}$ \\
\bottomrule
    \end{tabular}
    }
    \caption{NLI average accuracy results across 5 seeds using DistilMBERT by freezing strategy, language, aligner, and filtering threshold. Aligner names: FA - FastAlign, AA - AWESOME-align, BD - Bilingual Dictionary. The highest average accuracy value for each language is highlighted in bold.}
    \label{table:results_distilMBERT_NLI_filtering}
    \end{table*}
    

%% file: large_tables/pos-distilmbert-fa-filtering.tex
    \begin{table*}[ht]
    \centering
    \adjustbox{max width=\linewidth}{
    \begin{tabular}{lccccccccccccc}
    \toprule
     & \textbf{FT Only} & \multicolumn{6}{c}{\textbf{vanilla realignment}} & \multicolumn{6}{c}{\textbf{\textsc{AlignFreeze} with front-freezing}}\\ 
    \cline{2-14}
     & -  & 0\% & 25\% & 37\% & 50\% & 62\% & 75\% & 0\% & 25\% & 37\% & 50\% & 62\% & 75\%\\
    \midrule
    Afrikaans & 85.5$_{\pm 0.2}$ & 86.4$_{\pm 0.3}$ & 86.3$_{\pm 0.2}$ & 86.4$_{\pm 0.2}$ & 86.4$_{\pm 0.2}$ & 86.2$_{\pm 0.5}$ & \textbf{86.6}$_{\pm 0.3}$ & 86.2$_{\pm 0.2}$ & 86.2$_{\pm 0.2}$ & 86.3$_{\pm 0.2}$ & 86.3$_{\pm 0.4}$ & 86.2$_{\pm 0.2}$ & 86.2$_{\pm 0.2}$ \\
Arabic & 51.7$_{\pm 1.7}$ & 63.9$_{\pm 0.5}$ & 63.6$_{\pm 0.6}$ & 63.9$_{\pm 0.9}$ & 63.6$_{\pm 0.8}$ & 64.1$_{\pm 0.9}$ & 63.6$_{\pm 0.8}$ & 63.3$_{\pm 0.5}$ & 63.5$_{\pm 0.7}$ & 63.9$_{\pm 0.2}$ & 63.8$_{\pm 0.7}$ & \textbf{64.2}$_{\pm 0.5}$ & 64.0$_{\pm 0.4}$ \\
Bulgarian & 85.0$_{\pm 0.5}$ & \textbf{87.4}$_{\pm 0.2}$ & 87.2$_{\pm 0.2}$ & 87.3$_{\pm 0.3}$ & 87.3$_{\pm 0.3}$ & 87.1$_{\pm 0.4}$ & 87.1$_{\pm 0.2}$ & 87.1$_{\pm 0.3}$ & 87.1$_{\pm 0.3}$ & 87.1$_{\pm 0.2}$ & 87.1$_{\pm 0.1}$ & 87.0$_{\pm 0.4}$ & 87.1$_{\pm 0.2}$ \\
Catalan & 86.6$_{\pm 0.4}$ & \textbf{87.8}$_{\pm 0.2}$ & \textbf{87.8}$_{\pm 0.2}$ & 87.7$_{\pm 0.1}$ & \textbf{87.8}$_{\pm 0.2}$ & 87.7$_{\pm 0.1}$ & 87.7$_{\pm 0.2}$ & 87.6$_{\pm 0.3}$ & 87.7$_{\pm 0.2}$ & 87.7$_{\pm 0.3}$ & \textbf{87.8}$_{\pm 0.1}$ & 87.7$_{\pm 0.1}$ & 87.6$_{\pm 0.1}$ \\
Chinese & 64.3$_{\pm 1.4}$ & 66.2$_{\pm 0.5}$ & 66.4$_{\pm 0.7}$ & \textbf{66.7}$_{\pm 0.4}$ & 66.4$_{\pm 0.6}$ & 66.5$_{\pm 0.3}$ & 66.4$_{\pm 0.2}$ & 66.6$_{\pm 0.5}$ & 66.5$_{\pm 0.5}$ & 66.6$_{\pm 0.3}$ & 66.6$_{\pm 0.4}$ & 66.6$_{\pm 0.4}$ & 66.5$_{\pm 0.3}$ \\
Czech & 79.1$_{\pm 0.7}$ & \textbf{84.6}$_{\pm 0.3}$ & 84.2$_{\pm 0.2}$ & 84.3$_{\pm 0.2}$ & 84.2$_{\pm 0.3}$ & 84.1$_{\pm 0.4}$ & 84.1$_{\pm 0.3}$ & 83.7$_{\pm 0.3}$ & 83.7$_{\pm 0.2}$ & 83.8$_{\pm 0.1}$ & 83.8$_{\pm 0.3}$ & 83.7$_{\pm 0.3}$ & 83.8$_{\pm 0.4}$ \\
\bottomrule
    \end{tabular}
    }
    \caption{PoS tagging average accuracy results across 5 seeds using DistilMBERT by freezing strategy, language, and filtering threshold. Aligner name: FA - FastAlign. The highest average accuracy value for each language is highlighted in bold.}
    \label{table:results_distilMBERT_PoS_filtering_fa}
    \end{table*}
    

%% file: large_tables/pos-distilmbert-aa-filtering.tex
    \begin{table*}[ht]
    \centering
    \adjustbox{max width=\linewidth}{
    \begin{tabular}{lccccccccccccc}
    \toprule
     & \textbf{FT Only} & \multicolumn{6}{c}{\textbf{vanilla realignment}} & \multicolumn{6}{c}{\textbf{\textsc{AlignFreeze} with front-freezing}}\\ 
    \cline{2-14}
     & -  & 0\% & 25\% & 37\% & 50\% & 62\% & 75\% & 0\% & 25\% & 37\% & 50\% & 62\% & 75\%\\
    \midrule
    Afrikaans & 85.5$_{\pm 0.2}$ & 86.4$_{\pm 0.3}$ & 86.5$_{\pm 0.3}$ & 86.6$_{\pm 0.2}$ & 86.5$_{\pm 0.2}$ & 86.6$_{\pm 0.3}$ & \textbf{86.7}$_{\pm 0.2}$ & 86.3$_{\pm 0.3}$ & 86.5$_{\pm 0.2}$ & 86.4$_{\pm 0.3}$ & 86.5$_{\pm 0.2}$ & 86.4$_{\pm 0.3}$ & 86.5$_{\pm 0.1}$ \\
Arabic & 51.7$_{\pm 1.7}$ & 63.6$_{\pm 0.3}$ & 63.3$_{\pm 0.2}$ & 63.7$_{\pm 0.6}$ & 63.3$_{\pm 0.5}$ & 63.7$_{\pm 0.7}$ & 63.6$_{\pm 0.7}$ & 63.0$_{\pm 0.5}$ & 63.4$_{\pm 0.4}$ & \textbf{63.8}$_{\pm 0.7}$ & 63.7$_{\pm 0.5}$ & 63.7$_{\pm 0.6}$ & \textbf{63.8}$_{\pm 0.5}$ \\
Bulgarian & 85.0$_{\pm 0.5}$ & \textbf{87.6}$_{\pm 0.3}$ & 87.4$_{\pm 0.4}$ & 87.3$_{\pm 0.3}$ & 87.3$_{\pm 0.2}$ & 87.2$_{\pm 0.3}$ & 87.2$_{\pm 0.2}$ & 87.3$_{\pm 0.2}$ & 87.3$_{\pm 0.3}$ & 87.3$_{\pm 0.2}$ & 87.3$_{\pm 0.3}$ & 87.2$_{\pm 0.2}$ & 87.3$_{\pm 0.2}$ \\
Catalan & 86.6$_{\pm 0.4}$ & \textbf{88.1}$_{\pm 0.2}$ & 88.0$_{\pm 0.2}$ & 87.8$_{\pm 0.1}$ & 87.9$_{\pm 0.1}$ & 87.9$_{\pm 0.1}$ & 87.8$_{\pm 0.2}$ & 87.8$_{\pm 0.2}$ & 87.8$_{\pm 0.2}$ & 87.6$_{\pm 0.1}$ & 87.6$_{\pm 0.1}$ & 87.7$_{\pm 0.2}$ & 87.7$_{\pm 0.2}$ \\
Chinese & 64.3$_{\pm 1.4}$ & 66.3$_{\pm 0.6}$ & \textbf{66.5}$_{\pm 0.4}$ & 66.3$_{\pm 0.3}$ & 66.2$_{\pm 0.4}$ & 66.2$_{\pm 0.5}$ & 66.0$_{\pm 0.5}$ & 66.3$_{\pm 0.4}$ & 66.4$_{\pm 0.5}$ & \textbf{66.5}$_{\pm 0.3}$ & 66.3$_{\pm 0.2}$ & 66.4$_{\pm 0.5}$ & 66.3$_{\pm 0.6}$ \\
Czech & 79.1$_{\pm 0.7}$ & \textbf{84.7}$_{\pm 0.4}$ & 84.5$_{\pm 0.3}$ & 84.6$_{\pm 0.4}$ & 84.5$_{\pm 0.3}$ & 84.5$_{\pm 0.2}$ & 84.5$_{\pm 0.3}$ & 84.0$_{\pm 0.2}$ & 84.1$_{\pm 0.2}$ & 84.0$_{\pm 0.3}$ & 84.0$_{\pm 0.2}$ & 84.1$_{\pm 0.1}$ & 84.0$_{\pm 0.1}$ \\
\bottomrule
    \end{tabular}
    }
    \caption{PoS tagging average accuracy results across 5 seeds using DistilMBERT by freezing strategy, language, and filtering threshold. Aligner name: AA - AWESOME-align. The highest average accuracy value for each language is highlighted in bold.}
    \label{table:results_distilMBERT_PoS_filtering_aa}
    \end{table*}
    

%% file: large_tables/pos-distilmbert-bd-filtering.tex
    \begin{table*}[ht]
    \centering
    \adjustbox{max width=\linewidth}{
    \begin{tabular}{lccccccccccccc}
    \toprule
     & \textbf{FT Only} & \multicolumn{6}{c}{\textbf{vanilla realignment}} & \multicolumn{6}{c}{\textbf{\textsc{AlignFreeze} with front-freezing}}\\ 
    \cline{2-14}
     & -  & 0\% & 25\% & 37\% & 50\% & 62\% & 75\% & 0\% & 25\% & 37\% & 50\% & 62\% & 75\%\\
    \midrule
    Afrikaans & 85.5$_{\pm 0.2}$ & 85.6$_{\pm 0.4}$ & 86.1$_{\pm 0.3}$ & 85.8$_{\pm 0.4}$ & 85.7$_{\pm 0.4}$ & 85.7$_{\pm 0.4}$ & 85.6$_{\pm 0.2}$ & 86.1$_{\pm 0.3}$ & \textbf{86.3}$_{\pm 0.3}$ & 86.0$_{\pm 0.2}$ & 86.2$_{\pm 0.1}$ & \textbf{86.3}$_{\pm 0.2}$ & 86.1$_{\pm 0.3}$ \\
Arabic & 51.7$_{\pm 1.7}$ & \textbf{66.6}$_{\pm 0.5}$ & 66.3$_{\pm 0.9}$ & 65.9$_{\pm 0.6}$ & 65.8$_{\pm 0.7}$ & 66.1$_{\pm 0.3}$ & 65.2$_{\pm 0.8}$ & 65.0$_{\pm 0.6}$ & 65.1$_{\pm 0.8}$ & 64.8$_{\pm 0.8}$ & 65.0$_{\pm 0.7}$ & 64.9$_{\pm 0.6}$ & 64.2$_{\pm 0.3}$ \\
Bulgarian & 85.0$_{\pm 0.5}$ & \textbf{87.6}$_{\pm 0.4}$ & \textbf{87.6}$_{\pm 0.4}$ & 87.4$_{\pm 0.3}$ & 87.5$_{\pm 0.2}$ & 87.4$_{\pm 0.1}$ & 87.2$_{\pm 0.3}$ & 87.2$_{\pm 0.3}$ & 87.2$_{\pm 0.3}$ & 87.2$_{\pm 0.3}$ & 87.2$_{\pm 0.4}$ & 87.1$_{\pm 0.3}$ & 87.0$_{\pm 0.3}$ \\
Catalan & 86.6$_{\pm 0.4}$ & 88.4$_{\pm 0.1}$ & 88.4$_{\pm 0.2}$ & \textbf{88.5}$_{\pm 0.1}$ & 88.4$_{\pm 0.1}$ & 88.4$_{\pm 0.1}$ & 88.4$_{\pm 0.1}$ & 88.2$_{\pm 0.1}$ & 88.2$_{\pm 0.2}$ & 88.2$_{\pm 0.2}$ & 88.2$_{\pm 0.1}$ & 88.1$_{\pm 0.2}$ & 88.1$_{\pm 0.1}$ \\
Chinese & 64.3$_{\pm 1.4}$ & \textbf{67.4}$_{\pm 0.7}$ & 66.9$_{\pm 0.6}$ & \textbf{67.4}$_{\pm 0.6}$ & 67.0$_{\pm 0.5}$ & 67.3$_{\pm 0.9}$ & 66.9$_{\pm 0.8}$ & 67.3$_{\pm 0.6}$ & 67.1$_{\pm 0.4}$ & 67.3$_{\pm 0.5}$ & 67.2$_{\pm 0.5}$ & 67.2$_{\pm 0.7}$ & 67.1$_{\pm 0.7}$ \\
Czech & 79.1$_{\pm 0.7}$ & 85.3$_{\pm 0.5}$ & \textbf{85.4}$_{\pm 0.3}$ & 85.2$_{\pm 0.4}$ & 85.2$_{\pm 0.4}$ & 85.2$_{\pm 0.4}$ & 84.9$_{\pm 0.2}$ & 84.3$_{\pm 0.3}$ & 84.3$_{\pm 0.3}$ & 84.2$_{\pm 0.5}$ & 84.1$_{\pm 0.3}$ & 84.0$_{\pm 0.3}$ & 83.8$_{\pm 0.2}$ \\
\bottomrule
    \end{tabular}
    }
    \caption{PoS tagging average accuracy results across 5 seeds using DistilMBERT by freezing strategy, language, and filtering threshold. Aligner name: BD - Bilingual Dictionary. The highest average accuracy value for each language is highlighted in bold.}
    \label{table:results_distilMBERT_PoS_filtering_bd}
    \end{table*}

%% file: large_tables/pos-distilmbert-single-layer-realignment.tex
\begin{table*}[ht]
        \centering
        \adjustbox{max width=\linewidth}{
    \revision{
        \begin{tabular}{lcccccccc}
        \hline
         & FT Only & Layer 0 & Layer 1 & Layer 2 & Layer 3 & Layer 4 & Layer 5 & Layer 6 \\
        \hline
    Afrikaans & 85.5$_{\pm 0.2}$ & 85.7$_{\pm 0.3}$ & 85.6$_{\pm 0.3}$ & 85.6$_{\pm 0.3}$ & 85.4$_{\pm 0.3}$ & 85.7$_{\pm 0.3}$ & \textbf{85.7}$_{\pm 0.5}$ & 85.7$_{\pm 0.2}$ \\
Arabic & 51.7$_{\pm 1.5}$ & 51.3$_{\pm 1.0}$ & 60.4$_{\pm 0.8}$ & 62.2$_{\pm 0.7}$ & \textbf{62.2}$_{\pm 1.4}$ & 62.0$_{\pm 0.6}$ & 55.7$_{\pm 1.5}$ & 50.3$_{\pm 1.2}$ \\
Bulgarian & 85.0$_{\pm 0.4}$ & 85.9$_{\pm 0.4}$ & 86.7$_{\pm 0.4}$ & 86.8$_{\pm 0.2}$ & \textbf{86.9}$_{\pm 0.2}$ & 86.8$_{\pm 0.3}$ & 86.3$_{\pm 0.3}$ & 84.9$_{\pm 0.2}$ \\
Catalan & 86.6$_{\pm 0.4}$ & 87.3$_{\pm 0.4}$ & 87.6$_{\pm 0.4}$ & 87.6$_{\pm 0.3}$ & 87.9$_{\pm 0.2}$ & \textbf{87.9}$_{\pm 0.3}$ & 87.6$_{\pm 0.2}$ & 86.7$_{\pm 0.4}$ \\
Chinese & 64.3$_{\pm 1.2}$ & 65.5$_{\pm 0.7}$ & 66.6$_{\pm 0.5}$ & \textbf{66.7}$_{\pm 0.7}$ & 66.6$_{\pm 0.9}$ & 66.5$_{\pm 0.8}$ & 66.5$_{\pm 0.6}$ & 64.4$_{\pm 0.6}$ \\
Czech & 79.1$_{\pm 0.6}$ & 81.5$_{\pm 0.6}$ & \textbf{84.2}$_{\pm 0.4}$ & 84.0$_{\pm 0.2}$ & 83.8$_{\pm 0.3}$ & 83.2$_{\pm 0.3}$ & 82.3$_{\pm 0.2}$ & 79.5$_{\pm 0.3}$ \\
Danish & 87.8$_{\pm 0.3}$ & 87.7$_{\pm 0.4}$ & 88.0$_{\pm 0.4}$ & 87.9$_{\pm 0.3}$ & 88.0$_{\pm 0.3}$ & \textbf{88.4}$_{\pm 0.2}$ & 88.3$_{\pm 0.3}$ & 87.5$_{\pm 0.4}$ \\
\hline
        \end{tabular}
    }
        }
        \caption{\revision{PoS tagging average accuracy results across 5 seeds using distilMBERT when performing realignment while freezing all layers but one (Aligner: bilingual dictionary) }}
        \label{table:results_mBERT_PoS-single_layer_realignment}
        \end{table*}

%% file: large_tables/pos-distilmbert-single-layer-freezing.tex
\begin{table*}[ht]
        \centering
        \adjustbox{max width=\linewidth}{
        \revision{
        \begin{tabular}{lcccccccc}
        \hline
         & FT Only & Layer 0 & Layer 1 & Layer 2 & Layer 3 & Layer 4 & Layer 5 & Layer 6\\
        \hline
    Afrikaans & 85.5$_{\pm 0.2}$ & 85.7$_{\pm 0.2}$ & 85.8$_{\pm 0.2}$ & \textbf{85.9}$_{\pm 0.2}$ & 85.8$_{\pm 0.3}$ & 85.7$_{\pm 0.2}$ & 85.4$_{\pm 0.2}$ & 85.7$_{\pm 0.2}$ \\
Arabic & 51.7$_{\pm 1.5}$ & \textbf{66.7}$_{\pm 0.4}$ & 66.3$_{\pm 0.3}$ & 66.6$_{\pm 0.2}$ & 66.3$_{\pm 0.2}$ & 65.9$_{\pm 0.4}$ & 65.9$_{\pm 0.5}$ & 66.6$_{\pm 0.4}$ \\
Bulgarian & 85.0$_{\pm 0.4}$ & 87.6$_{\pm 0.3}$ & 87.5$_{\pm 0.2}$ & 87.6$_{\pm 0.3}$ & 87.6$_{\pm 0.2}$ & \textbf{87.7}$_{\pm 0.2}$ & 87.5$_{\pm 0.3}$ & 87.6$_{\pm 0.3}$ \\
Catalan & 86.6$_{\pm 0.4}$ & 88.4$_{\pm 0.1}$ & 88.4$_{\pm 0.1}$ & 88.3$_{\pm 0.1}$ & \textbf{88.4}$_{\pm 0.2}$ & 88.2$_{\pm 0.1}$ & 88.1$_{\pm 0.2}$ & 88.4$_{\pm 0.1}$ \\
Chinese & 64.3$_{\pm 1.2}$ & 67.4$_{\pm 0.7}$ & 67.3$_{\pm 0.5}$ & 67.4$_{\pm 0.7}$ & 67.2$_{\pm 0.8}$ & 67.2$_{\pm 0.6}$ & 66.7$_{\pm 0.6}$ & \textbf{67.5}$_{\pm 0.7}$ \\
Czech & 79.1$_{\pm 0.6}$ & 85.3$_{\pm 0.4}$ & 85.1$_{\pm 0.4}$ & \textbf{85.4}$_{\pm 0.4}$ & \textbf{85.4}$_{\pm 0.4}$ & 85.4$_{\pm 0.3}$ & 85.2$_{\pm 0.3}$ & \textbf{85.4}$_{\pm 0.4}$ \\
Danish & 87.8$_{\pm 0.3}$ & 88.3$_{\pm 0.2}$ & 88.3$_{\pm 0.2}$ & \textbf{88.4}$_{\pm 0.2}$ & \textbf{88.4}$_{\pm 0.2}$ & 88.1$_{\pm 0.2}$ & 88.2$_{\pm 0.2}$ & 88.3$_{\pm 0.2}$ \\
\hline
        \end{tabular}
        }
        }
        \caption{\revision{PoS tagging average accuracy results across 5 seeds using distilMBERT when performing realignment while freezing a single layer (Aligner: bilingual dictionary) }}
        \label{table:results_mBERT_PoS-single_layer_freezing}
        \end{table*}